\documentclass[lettersize,journal]{IEEEtran}

\usepackage{cite}
\usepackage{amsthm}
\usepackage{amsmath}
\usepackage{algorithm}
\usepackage{algorithmicx}
\usepackage{algpseudocode}
\usepackage{comment}
\usepackage{booktabs}
\usepackage{longtable}
\usepackage{stfloats}
\usepackage{bm}
\usepackage{setspace}
\usepackage[usenames,dvipsnames]{color}
\usepackage{amsfonts,amssymb,amsmath}
\usepackage{enumerate}
\usepackage{cases}
\usepackage{array}
\usepackage{colortbl}
\usepackage{slashbox}
\usepackage{multirow}
\usepackage{arydshln}
\usepackage{caption}
\usepackage{float}
\usepackage{xurl}
\usepackage[hidelinks]{hyperref}
\usepackage{multirow}
\usepackage{verbatim}
\usepackage{textcomp}
\usepackage{pifont}
\usepackage{graphicx}
\usepackage{epstopdf}
\usepackage{subcaption}
\DeclareMathOperator*{\argmax}{arg\,max}

\renewcommand\arraystretch{1.2}

\definecolor{Gray}{gray}{0.85}
\definecolor{mycyan}{cmyk}{.3,0,0,0}

\newcolumntype{a}{>{\columncolor{Gray}}c}
\newcolumntype{b}{>{\columncolor{white}}c}
\newcolumntype{d}{>{\columncolor{mycyan}}c}

\allowdisplaybreaks
\begin{document}
\captionsetup[figure]{name={Fig.},labelsep=period}
\title{Jointly Optimizing Memory, Computing and Offloading for Energy-Efficient DNN Inference}
\author{Yunchu~Han,~\IEEEmembership{Graduate Student Member,~IEEE,}
        Zhaojun~Nan,~\IEEEmembership{Member,~IEEE,}\\
        Sheng~Zhou,~\IEEEmembership{Senior Member,~IEEE,}
        and~Zhisheng Niu,~\IEEEmembership{Fellow,~IEEE}


\thanks{Yunchu Han, Sheng Zhou, and Zhisheng Niu are with the Beijing National Research Center for Information Science and Technology, Department of Electronic Engineering, Tsinghua University, Beijing 100084, China (e-mail: hyc23@mails.tsinghua.edu.cn; sheng.zhou@tsinghua.edu.cn; niuzhs@tsinghua.edu.cn).}
\thanks{Zhaojun Nan is with the School of Electronics and Internet of Things, Chongqing Polytechnic University of Electronic Technology, Chongqing 401331, China (e-mail: nanzhaojun@cquet.edu.cn).}
\thanks{An earlier version of this paper has been published in IEEE GLOBECOM 2025 \cite{Han1}.}

}

\maketitle


\vspace{-3em}
\begin{abstract}
In edge intelligence systems, computation offloading and dynamic voltage and frequency scaling (DVFS) are widely used to improve energy efficiency. However, as applications like deep neural network (DNN) inference evolve in complexity, their demands for memory resources have progressively increased, establishing memory as a critical bottleneck that impacts both latency and energy consumption. In this paper, we propose a joint optimization scheme integrating memory, computing and offloading for energy-efficient DNN inference. We formulate the energy minimization problem under deadline constraints as a mixed-integer nonlinear programming (MINLP) problem, where the main challenge arises from the coupling between offloading and resource optimization. To address this issue, we decompose the original problem into two subproblems under a fixed offloading policy. The local subproblem is transformed into an equivalent geometric programming (GP) problem to derive the optimal memory frequency and computing frequency, while the edge subproblem is solved via convex optimization. An efficient coordinate descent (CD) algorithm is further developed to jointly optimize all variables. Simulation results demonstrate that our proposed algorithm achieves near-optimal performance with the gap from the optimal policy being at most $9.3\%$, and reduces the energy consumption of devices by up to $16.1\%$ compared with benchmarks without memory frequency optimization.
\end{abstract}


\section{Introduction}
\IEEEPARstart {I}{n} the era of artificial intelligence (AI), deep neural networks (DNNs) have been extensively deployed on mobile devices for different applications, such as image recognition \cite{Yu2}, object detection \cite{Cao3}, AI-generated content (AIGC) services \cite{Du4}, etc. However, DNN inference usually requires substantial computation and memory access costs, leading to intolerable latency and enormous energy consumption. For example, using MobileNetv1 SSD for object detection on smartphones consumes over $20\%$ of the total energy consumption \cite{Wang5}. The inference process of the transformer-based language model produces an extremely high carbon emission footprint, which is even higher than that of training \cite{Wu6}. The model loading time of ResNet50 on OPPO R17 is about 1 second, while the inference time is about 1.3 seconds \cite{Luo7}. Therefore, it is significantly challenging to achieve real-time and energy-efficient DNN inference, especially on resource-limited devices. 

To address these challenges, wireless edge intelligence \cite{Sun8}, \cite{Wang9} has been proposed as a new paradigm, where mobile devices can offload their computational tasks to the edge node (e.g., base station, roadside unit) with the mobile edge computing (MEC) server so that the resource costs for devices can be reduced. In addition, the dynamic voltage and frequency scaling (DVFS) technique \cite{Rabaey10}, \cite{Han11} is widely applied to balance the latency and energy consumption by adjusting the frequency of processors (e.g., CPU frequency, GPU frequency). Combining these techniques, existing work has considered the joint optimization of computing resources and offloading to reduce the energy consumption of mobile devices from various perspectives (e.g., batch processing \cite{Shi12}, robust optimization \cite{Nan13}, integrated sensing-communication-computation \cite{Liu14}) and scenarios (e.g., Internet of Things (IoT) \cite{Panda15}, vehicular network \cite{Liu16}, unmanned aerial vehicle (UAV) \cite{Zhao17}). However, the above studies solely consider the joint optimization of computing and communication resources while neglecting the optimization of memory resources, which can significantly impact the latency and energy consumption of DNN inference \cite{Sze18}, \cite{Cai19}.

The demands for memory resources of DNNs have dramatically increased due to the proliferation of model parameters and the large volume of intermediate data. Memory resource optimization includes various approaches, such as memory footprint management and memory frequency adjustment. Many studies have focused on the memory footprint management with different methods (e.g., dynamic allocation of memory size for DNN inference \cite{Xie20}, efficient DNN block swapping to reduce the unnecessary memory copying and processing \cite{Wang21}, slicing and preload-aware memory planning to balance the latency and memory usage \cite{Li22}). In fact, DNN inference is increasingly constrained by the memory wall \cite{MemoryWall}, where the processing speed of computing units is often limited by the rate at which data is accessed from memory. Therefore, memory frequency has become a critical factor that impacts the inference time. Recently, a few studies have investigated the memory frequency adjustment \cite{Wang24}, \cite{Zhang25}. In \cite{Wang24}, the authors observe that increasing the memory frequency can accelerate some GPU kernels. In \cite{Zhang25}, it is observed that the adjustment of memory frequency can improve the performance of DNN inference on GPUs (i.e., latency per unit of energy consumption). However, the impact of jointly adjusting memory frequency and computing frequency on DNN inference time has not been well-understood.
\begin{figure}[t]
    \captionsetup{font={small}}
    \centering
    \includegraphics[width=0.48\textwidth]{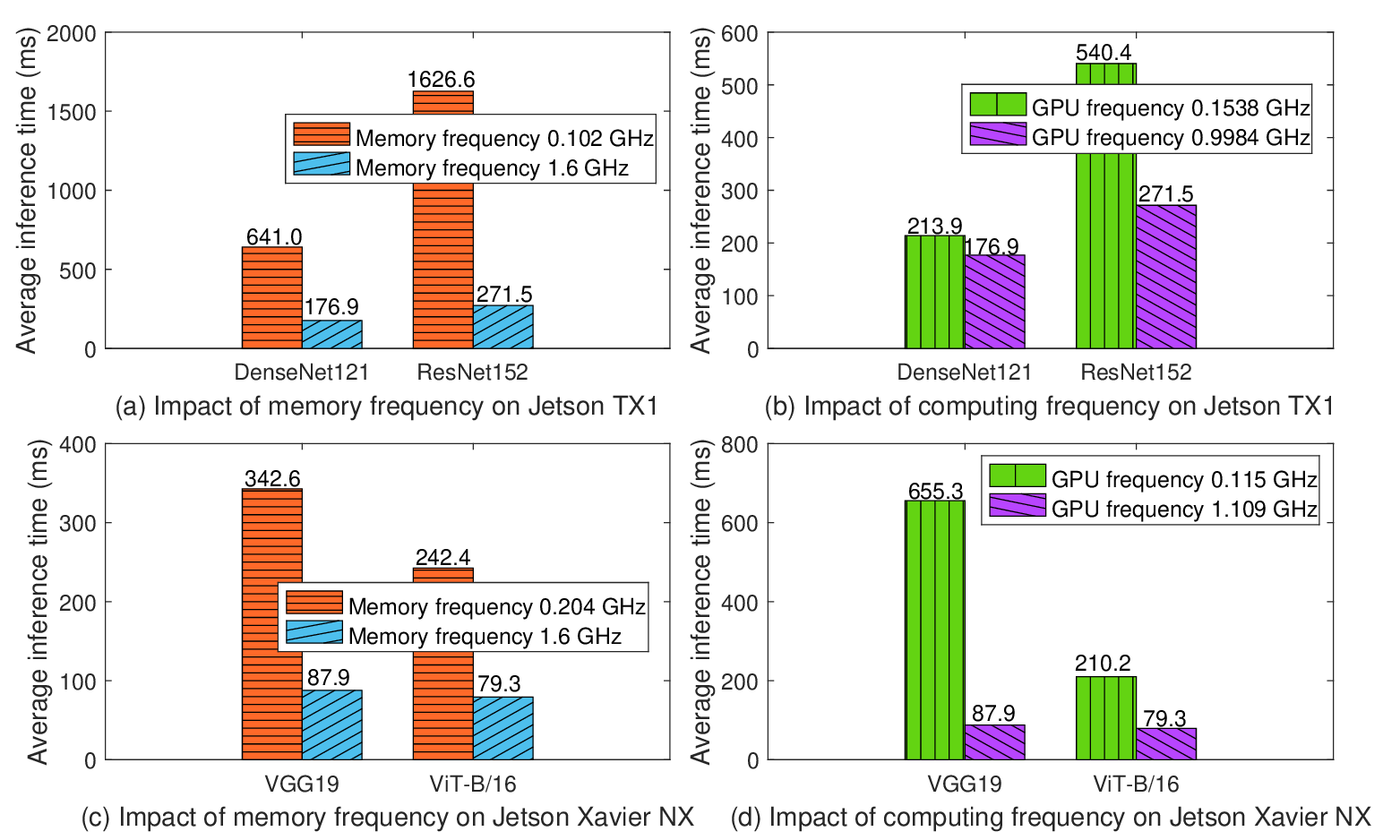}
    \caption{Average inference time of DNNs with different memory frequencies and computing frequencies on Jetson TX1 and Jetson Xavier NX.}\label{fig:Introduction}
    \vspace{-0.3cm}
\end{figure}

To further investigate the impact of memory frequency and computing frequency on DNN inference time, we measure the average inference time of four representative DNN models on the CIFAR-100 dataset \cite{Krizhevsky26} using two different devices with different frequency settings, as shown in Fig. \ref{fig:Introduction}. Increasing the memory frequency from 0.102 GHz to 1.6 GHz on Jetson TX1 reduces the inference time of DenseNet121 \cite{Huang27} and ResNet152 \cite{He28} by $72\%$ and $83\%$, respectively, which is considerably greater than the improvement achieved by increasing the computing frequency. A similar trend is observed on Jetson Xavier NX, where adjusting the memory frequency from 0.204 GHz to 1.6 GHz improves the inference time of VGG19 \cite{Simonyan29} and ViT-B/16 \cite{Dosovitskiy30} by $74\%$ and $67\%$, respectively. In fact, the impact of memory frequency and computing frequency on the inference time varies for different DNN models, reflecting their various computational and memory access costs. Therefore, jointly adjusting the memory frequency and computing frequency can provide more opportunities to reduce the energy consumption of DNN inference.  

In this paper, we investigate energy-efficient DNN inference in a multi-device edge intelligence system from a memory-computing-communication co-optimization perspective. Unlike most existing MEC-based studies that mainly optimize computing and communication resources, we explicitly consider the memory frequency of mobile devices as a controllable resource. For local inference, the memory frequency and computing frequency are jointly optimized to minimize the energy consumption. For edge inference, the uplink bandwidth allocation is optimized to minimize the energy consumption of transmission. Considering the combinatorial characteristic of binary offloading policies and resource optimization, we propose a coordinate descent (CD)-based algorithm to optimize the total energy consumption of all devices. The main contributions of this work are summarized as follows.
\begin{itemize}
\item
We consider a joint optimization framework for multi-device edge intelligence systems where mobile devices executing local inference jointly optimize the memory frequency and computing frequency to reduce the energy consumption. To this end, we characterize the impact of joint memory frequency and computing frequency scaling on DNN inference time. Specifically, the non-linear least square method is applied based on real data to fit the inference time as a function of both memory frequency and computing frequency. In addition, the power characteristics of memory frequency and computing frequency are investigated to obtain the energy consumption of DNN inference.
\item
We formulate an optimization problem that involves the joint optimization of memory frequency, computing frequency, uplink bandwidth and offloading policy to minimize the total energy consumption of all mobile devices, with the given deadline constraint. The problem is a mixed-integer nonlinear programming (MINLP) problem, and the strong coupling of offloading policy and resource allocation makes it difficult to directly obtain the optimal solution with acceptable complexity. 
\item
We develop an efficient decomposition-based solution algorithm by exploiting the structure of the problem. First, given the offloading policy, the problem is decomposed into a local subproblem optimizing the memory frequency and computing frequency, and an edge subproblem optimizing the uplink bandwidth. Then, the local subproblem is equivalently transformed into a GP problem, and the optimal solution can be derived. The optimal bandwidth allocation can be derived due to the convexity of the edge subproblem. Finally, we propose a CD-based algorithm to optimize the original offloading policy and obtain a local optimum for the original problem. 
\item 
Based on real data, we evaluate the performance of our proposed algorithm under different parameter settings. Specifically, we compare the energy consumption, inference time, memory frequency and computing frequency of devices among our proposed solution, memory-only solution and computing-only solution for local inference. In addition, our proposed algorithm achieves near-optimal performance for the overall problem and significantly outperforms the other benchmark policies. Finally, the impact of deadline and bandwidth on memory frequency and computing frequency in a specific case is given to exhibit the tradeoff between communication and memory/computing resources.
\end{itemize}

The remainder of this paper is organized as follows. Section \ref{sec:relatedwork} gives an overview of related work. Section \ref{sec:system} describes the system model and problem formulation. Section \ref{sec:Solution2} provides a detailed solution to the original problem. Section \ref{sec:Simulation} shows the simulation results to evaluate the proposed algorithm. Finally, Section \ref{sec:Conclusion} concludes this work.

\section{Related Work}\label{sec:relatedwork}
In this section, we review the existing literature on computation offloading, management and optimization of memory resources, and energy-efficient DNN inference.   

\subsection{Computation Offloading}
Extensive studies have focused on the computation offloading in MEC and edge intelligence systems, addressing how to allocate computing and communication resources and make offloading decisions. The binary offloading paradigm allows mobile devices to either offload tasks to the MEC server or execute local computation \cite{Nan31}, \cite{Chen32}. In \cite{Nan31}, a robust task offloading and resource allocation algorithm is proposed to address the imperfect computing capacity information. In \cite{Chen32}, a distributed online algorithm is proposed to minimize the energy consumption of ground devices in the aerial-based MEC system. To further reduce the energy consumption of computation and transmission, the partial offloading scheme combined with the DVFS technique has been studied \cite{Shi12}, \cite{Nan13}, \cite{Panda15}, \cite{Liu16}. In \cite{Shi12}, the authors consider an edge-device cooperation system, where devices optimize the computing frequency based on DVFS and offloading decisions while the edge server executes batch processing to minimize the total energy consumption. In \cite{Nan13}, the authors propose a robust DNN partitioning and resource allocation algorithm to reduce the energy consumption of mobile devices with probabilistic guarantees on deadlines. In \cite{Panda15}, a deep reinforcement learning (DRL) method is proposed to optimize the computation offloading ratio and computing frequency to minimize the energy consumption for real-time IoT applications. In \cite{Liu16}, a diffusion-based DRL algorithm is proposed to optimize the DNN partitioning policies, computation offloading and computing resource allocation in vehicular networks. However, these studies ignore the impact of memory resources on the latency and energy consumption of mobile devices. 

\subsection{Management and Optimization of Memory Resources}
A few studies have focused on memory footprint management and data-movement reduction \cite{Xie20}, \cite{Wang21}, \cite{Li22}, \cite{Dao33}, as well as memory frequency optimization \cite{Wang24}, \cite{Zhang25}, \cite{Guerreiro34}, \cite{Lu35}, \cite{Haj36}. Specifically, Smart-Zone \cite{Xie20} designs a memory-efficient management scheme for TrustZone-enabled IoT devices that dynamically optimizes secure and shared memory allocation to enable efficient and secure DNN inference. SwapNet \cite{Wang21} designs a zero-copy block swapping and model assembly mechanism to eliminate redundant memory operations for efficient DNN inference on edge devices with limited memory budgets. FlexNN \cite{Li22} adaptively manages and optimizes memory utilization to achieve latency-memory tradeoffs. FlashAttention \cite{Dao33} introduces an IO-aware attention with tiling and recomputation to reduce GPU memory accesses and latency. As for memory frequency, \cite{Wang24} characterizes a fine-grained model to estimate the memory access latency with memory and computing frequency scaling. \cite{Guerreiro34} decouples the GPU power consumption into computing and memory frequency domains, and presents models to predict the power consumption for various applications. \cite{Lu35} combines dynamic memory frequency scaling and DRAM low-power states to improve the energy consumption. \cite{Haj36} proposes a new multi-domain (e.g., compute, I/O, memory) power management technique to improve the energy efficiency of mobile System on Chip (SoC) with the fine-grained DVFS technique. However, it is difficult to directly apply the fine-grained models proposed by \cite{Wang24}, \cite{Guerreiro34} and \cite{Lu35} for memory frequency optimization in DNN inference. Although \cite{Zhang25} jointly optimizes DVFS and computation offloading ratio, the function relationship among the inference time, memory frequency and computing frequency is not characterized. 

\subsection{Energy-Efficient DNN Inference}
Recent studies have investigated energy-efficient DNN inference from different perspectives. For on-device or edge inference, E4 \cite{Zhang37} achieves energy-efficient DNN inference by integrating early exit with DVFS, where an attention-based module adaptively determines the DNN exit point and a profiler optimizes CPU/GPU frequencies for the layers before the selected exit point. Squeezer \cite{Wang38} focuses on efficient multi-DNN inference for edge video analytics through cross-model scheduling, where balanced operator grouping reduces scheduling complexity and a kernel scheduler coordinates parallel kernel execution to fully utilize the computing resources. For collaborative edge-cloud computing scenarios, \cite{Wang39} proposes a hybrid resource reserving and expansion scheme, where edge resources are managed locally and cloud resources are utilized for expansion to improve the system efficiency. \cite{Jiang40} combines early exit with DNN partitioning in device-edge collaboration, where DNN branch selection, partitioning point selection, computing and communication resource allocation are jointly optimized to reduce the energy consumption while satisfying the accuracy requirements. PIDC \cite{Jiang41} proposes a padding-aware framework that jointly optimizes DNN partitioning and padding data exchange among IoT devices to minimize inference latency. In addition, \cite{Zhang25} combines the memory frequency scaling with edge-cloud cooperation, and designs a DRL algorithm that co-optimizes DVFS and the offloading ratio to reduce the end-to-end latency and energy consumption.

Different from the existing work, this paper mainly investigates the impact of jointly adjusting memory frequency and computing frequency on DNN inference. We develop profiling-based models to characterize the impact of these two frequencies on both inference time and energy consumption. These models are incorporated into the proposed framework that jointly optimizes memory frequency, computing frequency, uplink bandwidth and offloading policy to minimize the total energy consumption of all mobile devices under the deadline constraint. Unlike the DRL-based method in \cite{Zhang25}, the proposed joint optimization scheme exploits the analytical structure of the formulated problem and provides tractable solutions to the decomposed subproblems. 

\section{System Model and Problem Formulation}\label{sec:system}
\subsection{Network Model}
\captionsetup[table]{labelsep=newline}
\begin{table*}[!htb]
    \footnotesize
    \caption{\label{tab:I}\textsc{Summary of Major Symbols}}
	\centering
	\begin{tabular}{p{1.0cm}<{\centering}|p{6.2cm}<{\centering}|p{1.0cm}<{\centering}|p{6.2cm}<{\centering}}
		\hline
        \textcolor{black}{\textbf{Symbol}} &\textcolor{black}{\textbf{Description} }
        & \textcolor{black}{\textbf{Symbol}} &\textcolor{black}{\textbf{Description} } \\
		\hline
		\textcolor{black}{$n$} &\textcolor{black}{Index of the $n$th mobile device}
		& \textcolor{black}{$\kappa_{n, \mathrm{m}}$} & \textcolor{black}{Memory energy coefficient of mobile device $n$}  \\
		\hline
		\textcolor{black}{$\mathcal{N}$} &\textcolor{black}{Set of $N$ mobile devices}
		& \textcolor{black}{$\kappa_{n, \mathrm{c}}$} &\textcolor{black}{Computing energy coefficient of mobile device $n$}  \\
		\hline
		\textcolor{black}{$f_{n, \mathrm{m}}$} &\textcolor{black}{Memory frequency of mobile device $n$}
		& \textcolor{black}{$f_{n, \min}$} &\textcolor{black}{Minimum memory frequency of mobile device $n$} \\
		\hline
		\textcolor{black}{$f_{n, \mathrm{c}}$} &\textcolor{black}{Computing frequency of mobile device $n$}
		& \textcolor{black}{$f_{n, \max}$} &\textcolor{black}{Maximum memory frequency of mobile device $n$} \\
		\hline
		\textcolor{black}{$t_{n}^{\mathrm{L}}$}& \textcolor{black}{Local inference time of mobile device $n$}
		& \textcolor{black}{$F_{n, \min}$} &\textcolor{black}{Minimum computing frequency of mobile device $n$} \\
		\hline
		\textcolor{black}{$e_{n}^{\mathrm{L}}$} &\textcolor{black}{Local energy consumption of mobile device $n$}
		& \textcolor{black}{$F_{n, \max}$} &\textcolor{black}{Maximum computing frequency of mobile device $n$} \\
		\hline
		\textcolor{black}{$t_{n}^{\mathrm{O}}$}& \textcolor{black}{Offloading time of mobile device $n$}
		& \textcolor{black}{$s_{n}$} &\textcolor{black}{Offloading data size of mobile device $n$} \\
		\hline
		\textcolor{black}{$e_{n}^{\mathrm{O}}$} &\textcolor{black}{Offloading energy consumption of mobile device $n$}
		& \textcolor{black}{$p_{n}$} &\textcolor{black}{Transmission power of mobile device $n$} \\
		\hline		
		\textcolor{black}{$t_{n}^{\mathrm{E}}$} &\textcolor{black}{Edge inference time of mobile device $n$}
		& \textcolor{black}{$h_{n}$} &\textcolor{black}{Channel gain of mobile device $n$} \\
		\hline
		\textcolor{black}{$D_n$} &\textcolor{black}{Inference task deadline of mobile device $n$}
	    & \textcolor{black}{$N_0$} &\textcolor{black}{Noise power spectral density} \\
		\hline
        \textcolor{black}{$B_n$} &\textcolor{black}{Uplink bandwidth of mobile device $n$}
        & \textcolor{black}{$B$} &\textcolor{black}{Total system bandwidth} \\
		\hline
        \hline
	\end{tabular}
    \vspace{-0.3cm}
\end{table*}

\begin{figure}[t]
  \captionsetup{font={small}}
  \centering
  \includegraphics[width=0.48\textwidth]{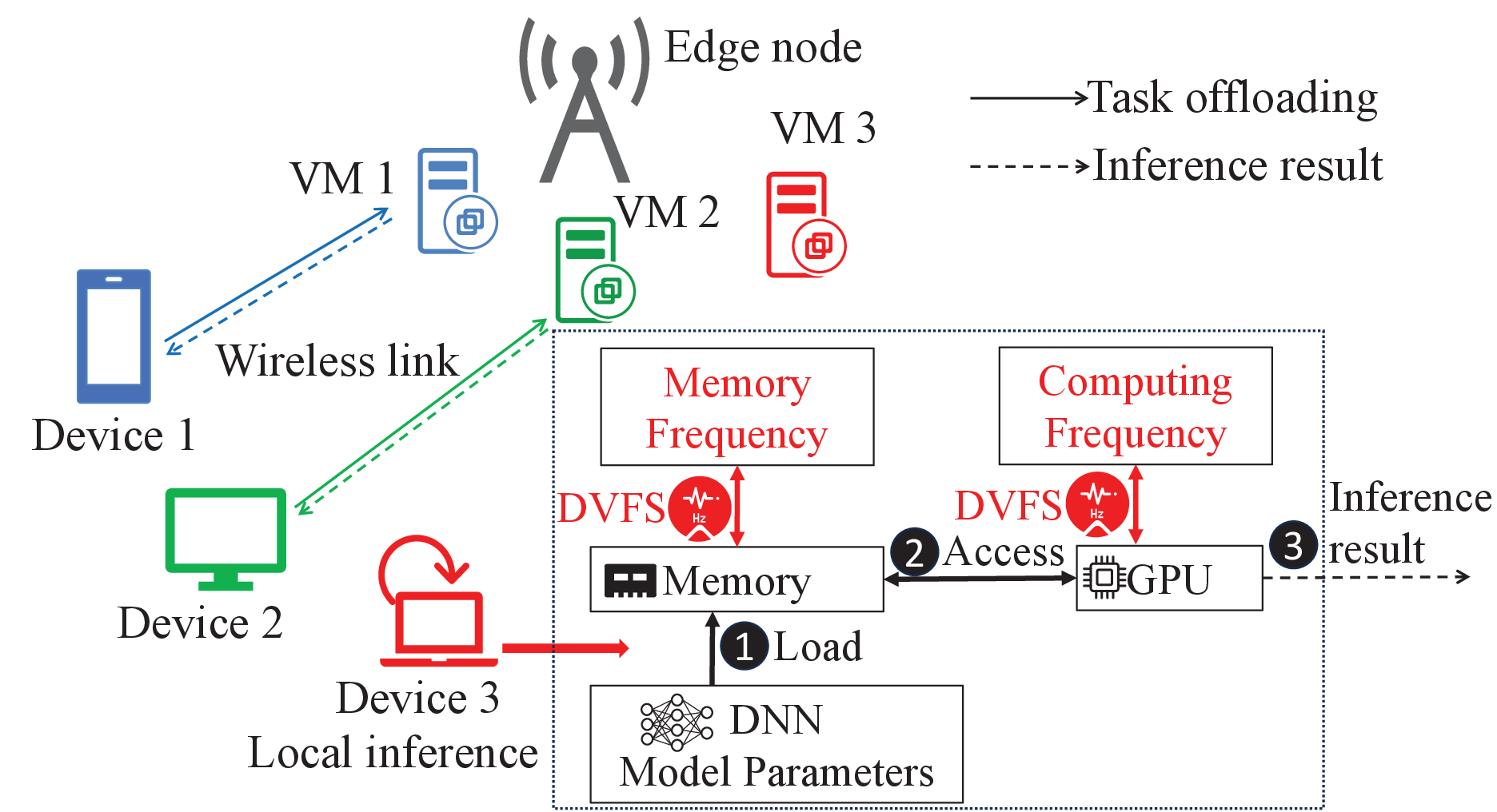}\\
  \caption{An illustration of the considered edge intelligence system with joint optimization of memory frequency and computing frequency.}
  \label{fig:SysMo}
  \vspace{-0.3cm}
\end{figure}

As shown in Fig. \ref{fig:SysMo}, we consider an edge intelligence system consisting of $N$ mobile devices, denoted by the set $\mathcal{N} \triangleq \{1, \ldots, N\}$, and a wireless edge node integrated with an MEC server. Each mobile device deploys pre-trained DNNs (e.g., ResNet \cite{He28}, VGG \cite{Simonyan29}, or ViT \cite{Dosovitskiy30}) to execute inference tasks (e.g., image recognition). We assume that the MEC server adopts a pre-configured resource allocation architecture, where an independent virtual machine (VM) is allocated to each device \cite{Liang42}. Each VM stores the corresponding DNN models and executes the inference task offloaded by the associated device in parallel. The computing and memory frequencies of each VM are set to the maximum values.\footnote{This work does not consider the computing resource allocation of the MEC server. However, our proposed method can be extended to scenarios that involve the optimization of computing resources for MEC servers.} The Frequency Division Multiple Access (FDMA) technique is adopted \cite{Lin43}, \cite{Zhao44}, where the interference between mobile devices can be ignored.\footnote{In this work, FDMA is adopted to obtain a tractable and continuous bandwidth allocation model, which facilitates the analysis of the overall optimization problem. However, our proposed optimization scheme can be extended to OFDMA-based systems by introducing the subcarrier allocation.} The binary offloading policy $x_n \in \{0, 1\}, \forall n \in \mathcal{N}$ is used, which means that each mobile device can process its inference task locally ($x_n = 0$) or offload its task to the corresponding VM for remote processing ($x_n = 1$). For example, device 1 and device 2 set to $x_n = 1$ offload their inference tasks to VM 1 and VM 2 with stronger processing capacities, while device 3 set to $x_n = 0$ completes its task locally. 

In this work, the joint optimization of memory frequency and computing frequency for mobile devices is considered. Specifically, the inference process can be generally divided into three steps. \ding{182}The DNN model parameters are first loaded into the memory (e.g., DRAM, SRAM). \ding{183}When performing the forward computation layer by layer, GPU kernels access the required input data and parameters from memory, perform the computation, and store the intermediate outputs back to memory as the inputs of subsequent layers. \ding{184}Finally, the forward computation is completed and the inference result is obtained. The memory frequency and computing frequency determine the memory access and forward computation time, respectively. Since both frequency domains contribute to the overall inference time and energy consumption, mobile devices executing local inference should jointly optimize these two frequencies to reduce the energy consumption. The major symbols used throughout this paper are summarized in Table \ref{tab:I}.
\begin{figure}[t]
    \captionsetup{font={small}}
    \centering
    \begin{subfigure}{0.48\linewidth}
        \centering
        \includegraphics[width=\linewidth]{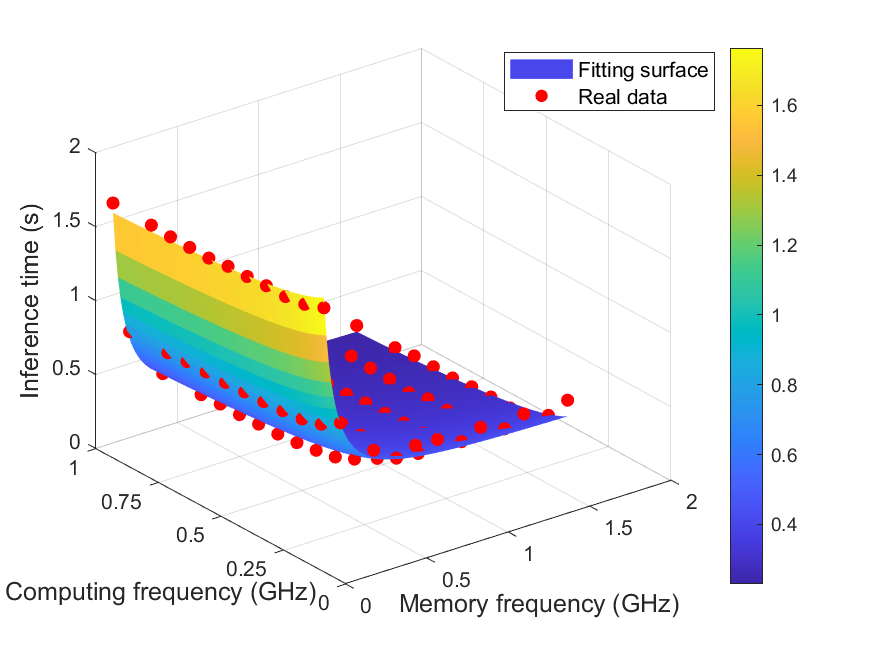}
        \caption{ResNet152 on Jetson TX1}
        \label{fig:ResNetTX1}
    \end{subfigure}
    \hfill
    \begin{subfigure}{0.48\linewidth}
        \centering
        \includegraphics[width=\linewidth]{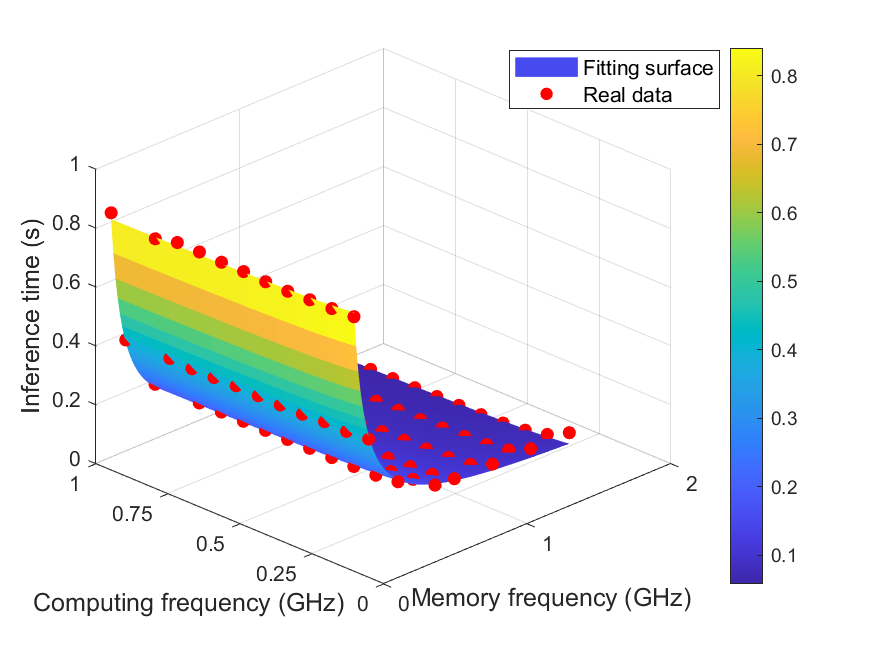}
        \caption{LSTM on Jetson TX1}
        \label{fig:LSTMTX1}
    \end{subfigure}
    \begin{subfigure}{0.48\linewidth}
        \centering
        \includegraphics[width=\linewidth]{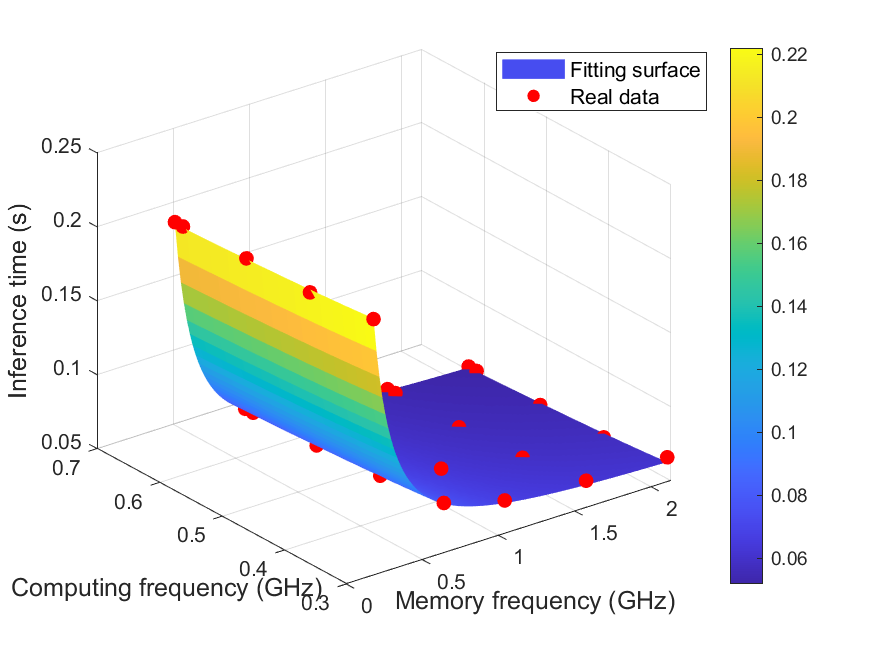}
        \caption{ViT-B/16 on Jetson Orin Nano}
        \label{fig:ViTOrinNano}
    \end{subfigure}
    \hfill
    \begin{subfigure}{0.48\linewidth}
        \centering
        \includegraphics[width=\linewidth]{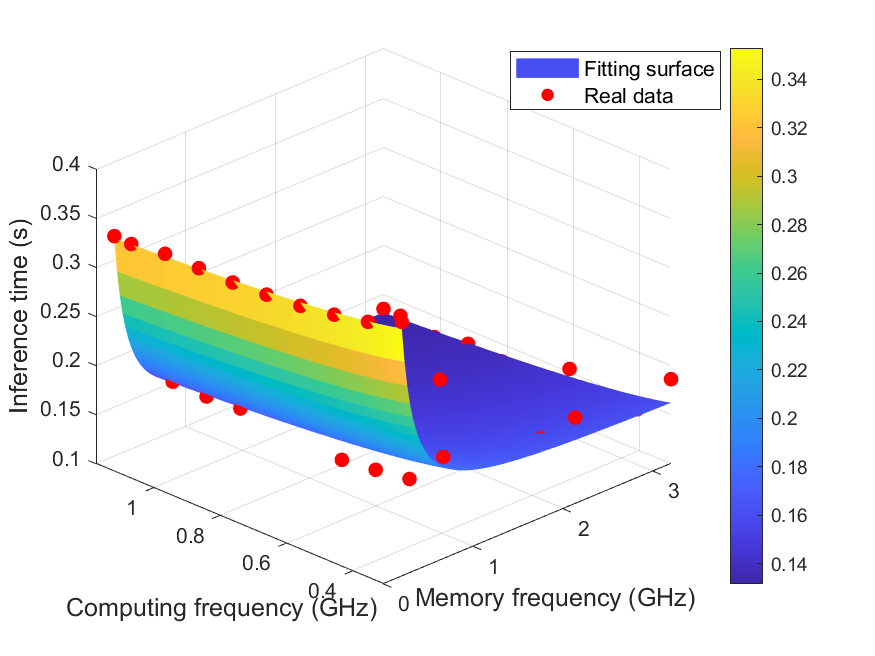}
        \caption{Swin-B on Jetson Orin NX}
        \label{fig:SwinBOrinNX}
    \end{subfigure}
    \caption{Inference time of four representative DNNs under different combinations of memory frequency and computing frequency.}
    \label{fig:InferenceTime}
    \vspace{-0.3cm}
\end{figure}

\subsection{Inference Time and Energy Consumption}
In most existing studies, the inference time is modeled as the ratio between the required computation workload (e.g., CPU cycles or floating-point operations) and the processing capability \cite{Han11}, \cite{Nan45}. However, this formulation fails to account for the impact of memory frequency, which plays a crucial role in determining the inference time. To characterize the impact of joint memory frequency and computing frequency scaling, we measure the average inference time of several representative DNNs on different hardware platforms under various combinations of memory frequency and computing frequency, as shown in Fig. \ref{fig:InferenceTime}. Specifically, ResNet152 and Long Short-Term Memory (LSTM) \cite{Hochreiter46} are deployed on Jetson TX1 to execute inference using the CIFAR-100 and UCI Human Activity Recognition (UCI HAR) \cite{Anguita47} datasets, respectively. ViT-B/16 and Swin Transformer-Base (Swin-B) \cite{Liu48} are evaluated on Jetson Orin Nano and Jetson Orin NX using the CIFAR-100 and ImageNet \cite{Deng49} datasets, respectively. For each memory and computing frequency configuration, we perform 10 warm-up runs and calculate the average inference time over the same 100 samples. Based on the real data, we use the non-linear least square method to characterize the function relationship between inference time and these two frequencies, which is illustrated by the fitting surface. 
\begin{table}[t]
    \scriptsize
    \centering
    \caption{\textsc{Fitting parameters and accuracy}}
	\setlength{\tabcolsep}{2pt}
	\begin{tabular}{c|ccc|cc|cc}
		\hline
		\multirow{2}*{\textbf{Parameters}} & \multicolumn{3}{c|}{\textbf{Jetson TX1}} & \multicolumn{2}{c|}{\textbf{Jetson Orin Nano}} & \multicolumn{2}{c}{\textbf{Jetson Orin NX}} \\
		\cline{2-8}
		\multirow{2}*{} & \textbf{ResNet152} & \textbf{VGG19} & \textbf{LSTM} & \textbf{ResNet152} & \textbf{ViT-B/16} & \textbf{GNN} & \textbf{Swin-B} \\
		\hline
		$a_n$ & 0.039 & 0.110 & 0.051 & 0.012 & 0.012 & 0.060 & 0.020            \\
		\hline
		$b_n$ & 1.552 & 1.296 & 1.193 & 1.426 & 1.638 & 1.423 & 1.445            \\
		\hline
		$c_n$ & 0.211 & 0.099 & 0.029 & 0.112 & 0.043 & 0.081 & 0.131            \\
		\hline
		$d_n$ & 0.358 & 0.629 & 0.370 & 0.036 & 0.241 & 0.700 & 0.159            \\
		\hline
		$\mathrm{R}^2 \uparrow$ & 0.9917 & 0.9961 & 0.9989 & 0.9993 & 0.9996 & 0.9925 & 0.9791           \\
		\hline
		MSE ($\mathrm{s}^2$) $\downarrow$  & 0.0016 & 0.0017 & 0.0001 & 0.000002 & 0.000002 & 0.0004 & 0.0001 \\
		\hline
		RMSE (s) $\downarrow$ & 0.0396 & 0.0416 & 0.0080 & 0.0013 & 0.0014 & 0.0205 & 0.0115 \\
		\hline
		MAE (s) $\downarrow$ & 0.0315 & 0.0291 & 0.0055 & 0.0011 & 0.0011 & 0.0141 & 0.0088 \\
		\hline
		MAPE  $\downarrow$   & 7.65\% & 6.11\% & 4.00\% & 0.76\% & 1.36\% & 6.48\% & 5.38\% \\
		\hline 
	\end{tabular}
    \label{tab:fit}
	\vspace{-0.3cm}
\end{table}

Let $f_{n, \mathrm{m}}$ and $f_{n, \mathrm{c}}$ denote the memory frequency and computing frequency of mobile device $n$, respectively. Then, the local inference time of mobile device $n$ is characterized by
\begin{equation}
    t_n^{\mathrm{L}} = a_n f_{n, \mathrm{m}}^{-b_n} + c_n f_{n, \mathrm{c}}^{-d_n}, \forall n \in \mathcal{N},
\end{equation}
where $a_n$, $b_n$, $c_n$ and $d_n$ are positive coefficients relevant to the hardware, the type of DNNs, etc. The fitting parameters and accuracy metrics are summarized in Table \ref{tab:fit}. We analyze the mean squared error (MSE), root mean squared error (RMSE), mean absolute error (MAE), mean absolute percentage error (MAPE) and R-squared ($\mathrm{R}^2$) to provide a more comprehensive evaluation of the fitting accuracy. The fitted model achieves small MSE, RMSE, MAE and MAPE values, while $\mathrm{R}^2$ is larger than $0.99$ in most cases, which further indicates the effectiveness of this model.\footnote{Due to space limitations, we only present several specific cases in this paper. Experiment results for other DNNs are shown in \cite{Han1}, where the inference time exhibits a trend similar to that shown in Fig. \ref{fig:InferenceTime} and can be characterized by the same model form.} Parameters $a_n$ and $b_n$ reflect the impact of memory frequency, while $c_n$ and $d_n$ reflect the impact of computing frequency. It is observed that the inference time exhibits a significant decrease as the memory frequency increases, verifying the impact of memory frequency scaling. In addition, when the memory frequency is higher, there is a larger decrease in the inference time with the increase of computing frequency. However, with extremely low memory frequency, increasing the computing frequency cannot lead to a distinct effect. This is because during DNN inference, GPUs not only perform arithmetic operations but also frequently access model parameters, input/output feature maps and intermediate activations. When the memory frequency is low, memory access becomes the major bottleneck, and the computing units may stall while waiting for the required data. In this case, increasing the computing frequency alone cannot effectively reduce the inference latency. Optimizing only the computing frequency while keeping the memory frequency fixed may lead to a suboptimal energy-latency tradeoff, especially when the DNN inference workload is memory-bound. This observation further supports the motivation of this work, and it is necessary to jointly consider the memory frequency and computing frequency in the proposed optimization framework. 
\begin{figure}[t]
  \centering
  \captionsetup{font={small}}
  \includegraphics[width=0.45\textwidth]{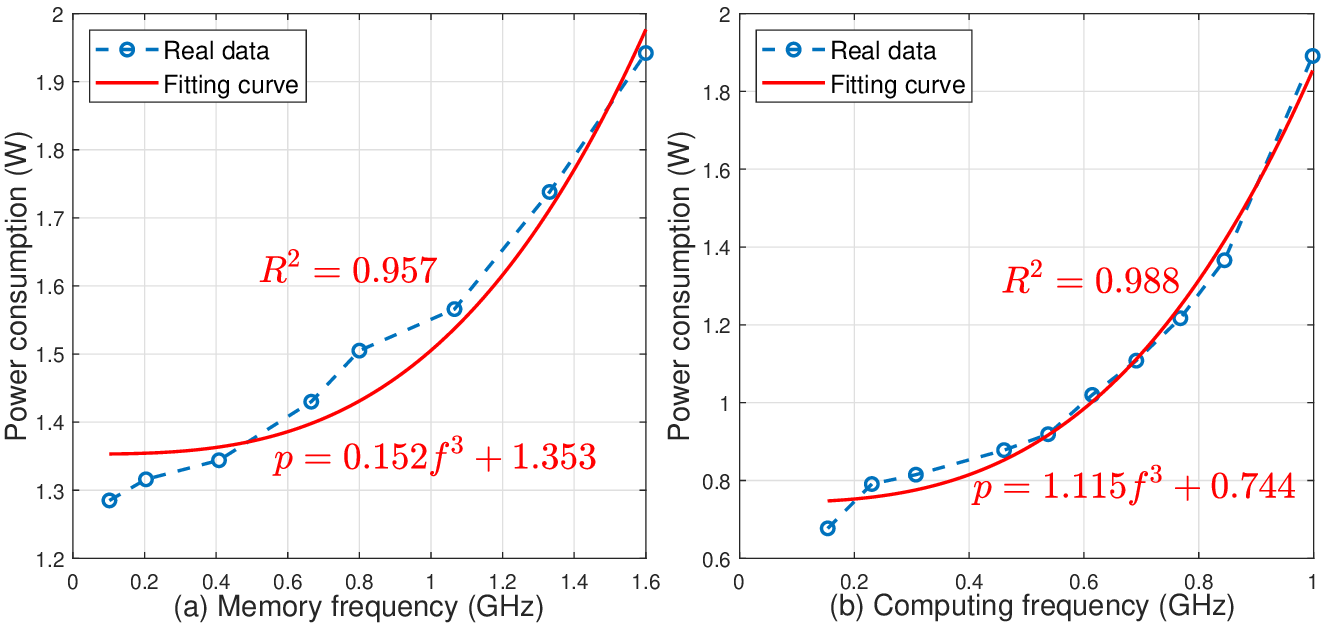}\\
  \caption{Power consumption characteristics of memory frequency and computing frequency scaling on Jetson TX1.}
  \label{fig:PowerTX1}
  \vspace{-0.3cm}
\end{figure}
\begin{figure}[t]
  \centering
  \captionsetup{font={small}}
  \includegraphics[width=0.45\textwidth]{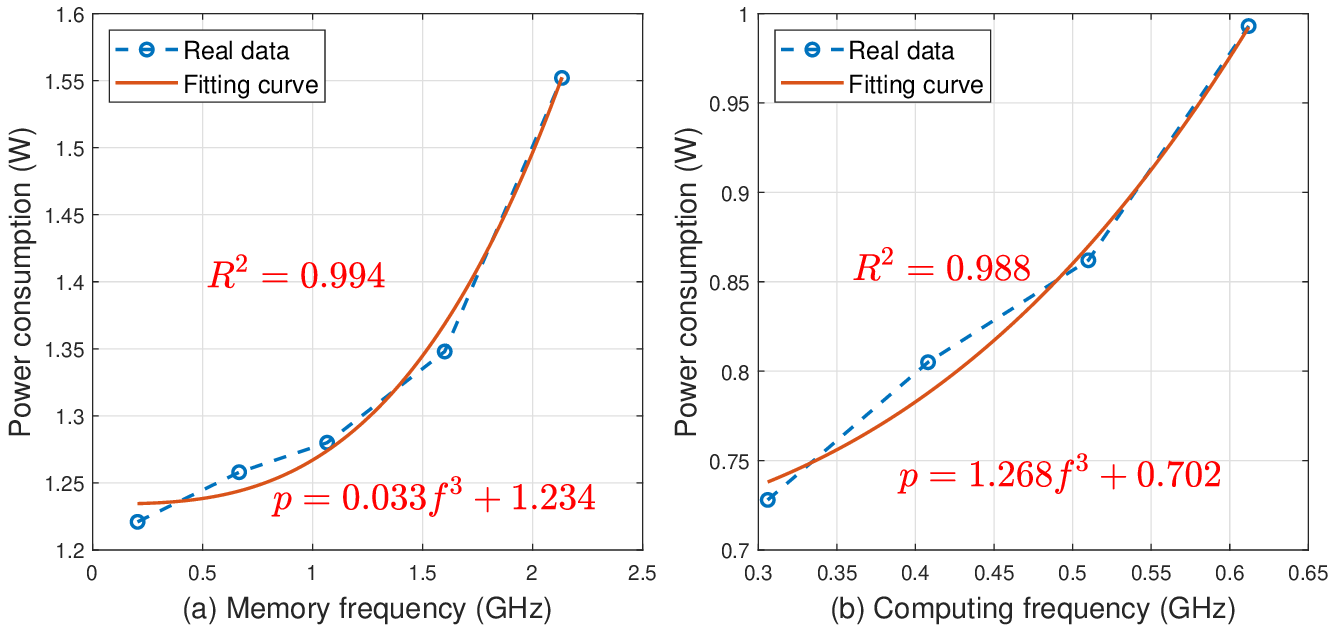}\\
  \caption{Power consumption characteristics of memory frequency and computing frequency scaling on Jetson Orin Nano.}
  \label{fig:PowerOrin}
  \vspace{-0.3cm}
\end{figure}

Based on the above analysis, the local inference time is formulated as a function related to both memory frequency and computing frequency. Moreover, it is essential to investigate the power characteristics associated with these two frequencies in order to accurately characterize the overall energy consumption. The dynamic power consumption is typically modeled as $P = \kappa f^3$, where $\kappa$ and $f$ denote the energy coefficient and computing frequency, respectively \cite{Yahya50}, \cite{Zeng51}. However, the total power consumption should account for not only the computing frequency domain, but also the memory frequency domain \cite{Guerreiro34}. To this end, we obtain the power consumption under different settings of memory frequency and computing frequency on two devices using the NVIDIA Tegrastats toolkit \cite{NVIDIA52}, as illustrated in Fig. \ref{fig:PowerTX1} and Fig. \ref{fig:PowerOrin}. Denote $\kappa_{n, \mathrm{m}}$ and $\kappa_{n, \mathrm{c}}$ as the energy coefficients for memory frequency and computing frequency, respectively. Then, the power consumption of mobile device $n$ is modeled as
\begin{equation}
    p_n^{\mathrm{L}} = \kappa_{n, \mathrm{m}} f_{n, \mathrm{m}}^3 + \kappa_{n, \mathrm{c}} f_{n, \mathrm{c}}^3 + \sigma_n, \forall n \in \mathcal{N},
\end{equation}
where $\sigma_n$ denotes the constant term representing the static power.\footnote{In this work, we fit the measured power consumption using the cubic-form model, since it has been widely adopted as a tractable approximation of dynamic power consumption. However, the proposed optimization framework and method do not rely on a strict cubic power model.} The mean squared errors (MSEs) of the fitting results for the power consumption of memory frequency and computing frequency on Jetson TX1 are $2.0 \times 10^{-3} \, \mathrm{W}^2$ and $1.4 \times 10^{-3} \, \mathrm{W}^2$, respectively. For Jetson Orin Nano, the corresponding MSEs are $1.3 \times 10^{-4} \, \mathrm{W}^2$ and $1.1 \times 10^{-4} \, \mathrm{W}^2$, respectively. The MSEs are extremely small and $\mathrm{R}^2$ is larger than $0.95$, verifying the effectiveness of our fitted power model. In addition, it is observed that the static power of memory is larger than that of computing, while the dynamic power presents an opposite result. The different latency and power consumption characteristics of memory and computing frequency directly determine the energy consumption of local inference for mobile device $n$, which is given by
\begin{equation}
    e_n^{\mathrm{L}} = p_n^{\mathrm{L}} t_{n}^{\mathrm{L}}, \forall n \in \mathcal{N}.
\end{equation}

For mobile devices executing edge inference, the energy consumption of transmission should be considered. Denote $B_n$ as the uplink bandwidth allocated to mobile device $n$ for offloading the inference task, where $B_n$ is constrained by the total bandwidth $B$ (i.e., $\sum_{n\in\mathcal{N}} B_n \leq B, B_n \geq 0$). Then, the spectral efficiency of the wireless link between the edge node and mobile device $n$ is given by $\eta_n = \log_2(1 + \frac{p_n h_n}{B_n N_0})$, where $p_n$ is the transmission power of device $n$, $h_n$ is the channel gain between device $n$ and the edge node, and $N_0$ is the power spectral density of the noise. Therefore, the transmission time of device $n$ for offloading its task to the edge node is 
\begin{equation}
    t_n^{\mathrm{O}} = \frac{s_n}{B_n \eta_n}, \forall n \in \mathcal{N},
\end{equation}
where $s_n$ is the input data size of the inference task for mobile device $n$. Correspondingly, the transmission energy consumption of mobile device $n$ is given by
\begin{equation}
    e_n^{\mathrm{O}} = \frac{p_n s_n}{B_n \eta_n}, \forall n \in \mathcal{N}.
\end{equation}

Based on the assumptions in the last subsection, for a given DNN model and VM configuration, the edge inference time of device $n$ can be measured through offline profiling and treated as a constant, denoted by $t_n^{\mathrm{E}}$. Given that the size of the inference result is usually much smaller than the raw data and the edge node possesses significantly high transmission power and bandwidth, the result feedback delay from the edge node to mobile devices is ignored \cite{Nan31}, \cite{He53}.

\subsection{Problem Formulation}
Based on the above analysis, the energy consumption of mobile device $n$ is given by
\begin{equation}
E_n= \left(1 - x_{n}\right)e_{n}^{\mathrm{L}} + x_n e_{n}^{\mathrm{O}}, \forall n \in \mathcal{N},
\end{equation}
and the corresponding inference time of mobile device $n$ is
\begin{equation}
T_n = \left(1 - x_n\right) t_n^{\mathrm{L}} + x_n \left(t_n^{\mathrm{O}} + t_n^{\mathrm{E}}\right), \forall n \in \mathcal{N}.
\end{equation}
Due to the real-time demand for DNN inference, the inference time should satisfy the deadline constraint, i.e., $T_n \leq D_n, \forall n \in \mathcal{N},$ where $D_n$ is the inference task deadline of mobile device $n$. The objective is to minimize the total energy consumption of mobile devices by jointly optimizing the binary offloading policy $\mathbf{x} \triangleq \left\{ x_n \right\}_{n \in \mathcal{N}}$, memory frequency $\mathbf{f_{M}} \triangleq \left\{ f_{n, \mathrm{m}} \right\}_{n \in \mathcal{N}}$, computing frequency $\mathbf{f_{C}} \triangleq \left\{ f_{n, \mathrm{c}} \right\}_{n \in \mathcal{N}}$, and uplink bandwidth $\mathbf{b} \triangleq \left\{ B_n \right\}_{n \in \mathcal{N}}$, while the deadline constraint should be satisfied. Therefore, the optimization problem is formulated as
\begin{subequations}\label{eq:Problem1}
\begin{alignat}{2}
& \min _{\mathbf{x}, \mathbf{b}, \mathbf{f_M}, \mathbf{f_C}} \sum_{n \in \mathcal{N}} E_n \label{eq:Problem1a} \\
& \ \text {s.t.} \, T_n \leq D_n, \forall n \in \mathcal{N}, \label{eq:Problem1b} \\
& \ \quad \ \, f_{n, \mathrm{m}} \in \left[f_{n, \min}, f_{n, \max}\right], \forall n \in \mathcal{N}, \label{eq:Problem1c} \\
& \ \quad \ \, f_{n, \mathrm{c}} \in \left[F_{n, \min}, F_{n, \max}\right], \forall n \in \mathcal{N}, \label{eq:Problem1d} \\
& \ \quad \ \, \sum_{n \in \mathcal{N}} x_n B_n \leq B, \label{eq:Problem1e} \\
& \ \quad \ \, x_n \in \{0, 1\}, \forall n \in \mathcal{N}, \label{eq:Problem1f} \\
& \ \quad \ \, B_n \geq 0, \forall n \in \mathcal{N}, \label{eq:Problem1g}
\end{alignat}
\end{subequations}
where (\ref{eq:Problem1b}) represents the deadline constraint of inference time, (\ref{eq:Problem1c}) and (\ref{eq:Problem1d}) correspond to the memory frequency and computing frequency constraints, (\ref{eq:Problem1e}) and (\ref{eq:Problem1g}) give the uplink bandwidth constraint, and (\ref{eq:Problem1f}) indicates the offloading policy constraint. The objective function is generally non-convex, and the introduction of memory frequency increases the complexity of problem (\ref{eq:Problem1}). In addition, even when the memory frequency is given, problem (\ref{eq:Problem1}) is still an MINLP problem, which is generally NP-hard. To solve this problem, we first assume that the offloading policy is given and derive the optimal solutions to the two subproblems. Then, we propose a low-complexity algorithm to solve the original problem. The detailed solutions are given in Section \ref{sec:Solution2}.

\section{Joint Optimization of Memory Frequency, Computing Frequency and Offloading}\label{sec:Solution2}
\captionsetup{font={small}}
\begin{figure*}[t]
\centering
\includegraphics[width=0.85\textwidth]{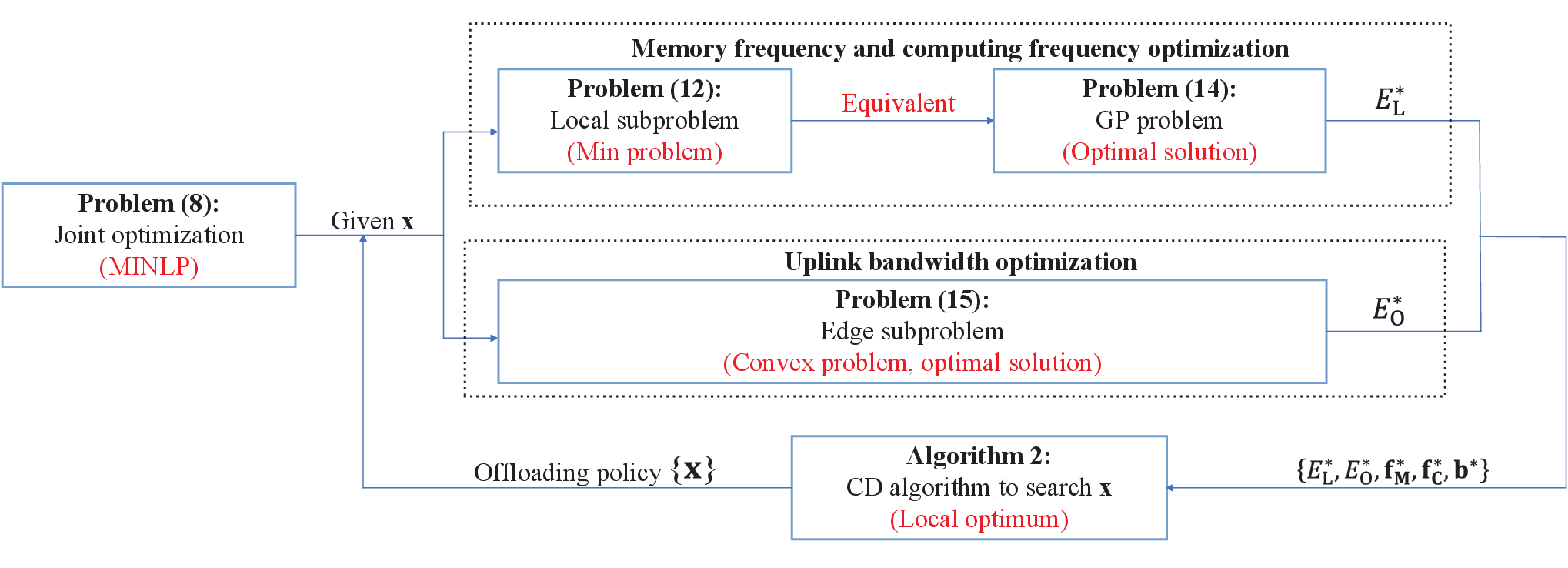}\\
\caption{\leftline{The overview of the proposed optimization problems and corresponding solving methods.}}
\label{fig:Overview}
\vspace{-0.3cm}
\end{figure*}
In this section, we use a decoupling method to solve the optimization problem (\ref{eq:Problem1}). We first assume that the offloading policy is given, and derive two independent subproblems to optimize the energy consumption of local inference and task offloading. Then, the local subproblem (i.e., optimization of memory frequency and computing frequency) is transformed into an equivalent GP problem, and the optimal solution is obtained. Meanwhile, since the edge subproblem (i.e., uplink bandwidth allocation) is a convex optimization problem, the optimal solution can be obtained. Finally, an efficient CD algorithm is proposed to optimize the offloading policy.

\subsection{Problem Decomposition and Overview}
Problem (\ref{eq:Problem1}) can be decomposed into independent subproblems with separate optimization variables and constraints, which are relatively easy to solve. First, with the given offloading policy, the local subproblem is given as follows
\begin{subequations}\label{eq:Problem2}
\begin{alignat}{2}
    & \min _{\mathbf{f_M}, \mathbf{f_C}} E(\mathbf{f_M}, \mathbf{f_C} \mid \mathbf{x}) \\
    & \ \text{s.t.} \, (\mathrm{\ref{eq:Problem1b}}), (\mathrm{\ref{eq:Problem1c}}), (\mathrm{\ref{eq:Problem1d}}), 
\end{alignat}
\end{subequations}
where $E(\mathbf{f_M}, \mathbf{f_C} \mid \mathbf{x})$ is the objective function corresponding to the total energy consumption of all mobile devices that execute local inference. Meanwhile, with the given offloading policy, the edge subproblem is expressed as
\begin{subequations}\label{eq:Problem3}
\begin{alignat}{2}
    & \min _{\mathbf{b}} E(\mathbf{b} \mid \mathbf{x}) \\
    & \ \text{s.t.} \, (\mathrm{\ref{eq:Problem1b}}), (\mathrm{\ref{eq:Problem1e}}), (\mathrm{\ref{eq:Problem1g}}), 
\end{alignat}
\end{subequations}
where $E(\mathbf{b} \mid \mathbf{x})$ is the objective function corresponding to the total energy consumption of all mobile devices that offload their inference tasks. Furthermore, with the given memory frequency, computing frequency and uplink bandwidth, the task offloading subproblem is written as
\begin{subequations}\label{eq:Problem4}
\begin{alignat}{2}
    & \min _{\mathbf{x}} E(\mathbf{x} \mid \mathbf{f_M}, \mathbf{f_C}, \mathbf{b}) \\
    & \ \text{s.t.} \, (\mathrm{\ref{eq:Problem1b}}), (\mathrm{\ref{eq:Problem1e}}), (\mathrm{\ref{eq:Problem1f}}),
\end{alignat}
\end{subequations}
where $E(\mathbf{x} \mid \mathbf{f_M}, \mathbf{f_C}, \mathbf{b})$ is the objective function corresponding to the total energy consumption of all mobile devices. We employ the Tammer decomposition method \cite{Tammer54} to decompose the original problem into tractable subproblems and solve them alternately. In the remainder of this section, we will provide specific solutions and analysis for the optimization of memory frequency and computing frequency, uplink bandwidth allocation and the task offloading subproblems. The overview of the decomposition and solutions is illustrated in Fig. \ref{fig:Overview}. 

\subsection{Optimization of Memory Frequency and Computing Frequency}
The set of all mobile devices is denoted by $\mathcal{N}$. When the offloading policy $\mathbf{x} \triangleq \{ x_n \}_{n \in \mathcal{N}}$ is given, we can divide the set $\mathcal{N}$ into two subsets $\mathcal{N}_0$ and $\mathcal{N}_1 = \mathcal{N} \setminus \mathcal{N}_0$, where $\mathcal{N}_0 \triangleq \{ n \in \mathcal{N} \big| x_n = 0 \}$ represents the set of devices executing local inference, and $\mathcal{N}_1 \triangleq \{ n \in \mathcal{N} \big| x_n = 1\}$ represents the set of devices offloading the inference tasks to the edge node. The local subproblem is to minimize the total energy consumption of devices with the deadline constraint by optimizing the memory frequency and computing frequency, which is formulated as
\begin{subequations}\label{eq:Problem5}
\begin{alignat}{2}
    & E_{\mathrm{L}}^{\ast} \triangleq \min _{\mathbf{f_M}, \mathbf{f_C}} \sum_{n \in \mathcal{N}_0} \left( \kappa_{n, \mathrm{m}} f_{n, \mathrm{m}}^3 + \kappa_{n, \mathrm{c}} f_{n, \mathrm{c}}^3 + \sigma_n \right) t_n \label{eq:Problem5a}\\
    & \ \text{s.t.} \, t_n \leq D_n, \forall n \in \mathcal{N}_0, \label{eq:Problem5b} \\
    & \ \quad \ \, f_{n, \mathrm{m}} \in \left[f_{n, \min}, f_{n, \max}\right], \forall n \in \mathcal{N}_0, \label{eq:Problem5c} \\
    & \ \quad \ \, f_{n, \mathrm{c}} \in \left[F_{n, \min}, F_{n, \max}\right], \forall n \in \mathcal{N}_0, \label{eq:Problem5d}
\end{alignat}
\end{subequations}
where $t_n = a_n f_{n, \mathrm{m}}^{-b_n} + c_n f_{n, \mathrm{c}}^{-d_n}$ is the local inference time of mobile device $n$. To address this problem, we introduce an auxiliary variable $y > 0$ as the upper bound of the objective function (\ref{eq:Problem5a}), which is expressed as
\begin{equation}
    \sum_{n \in \mathcal{N}_0} \left( \kappa_{n, \mathrm{m}} f_{n, \mathrm{m}}^3 + \kappa_{n, \mathrm{c}} f_{n, \mathrm{c}}^3 + \sigma_n \right) t_n \leq y.
\end{equation}
Then, problem (\ref{eq:Problem5}) can be transformed into an equivalent problem, which is formulated as
\begin{subequations}\label{eq:Problem6}
\begin{alignat}{2}
    & \min _{\mathbf{f_M}, \mathbf{f_C}, y} y \label{eq:Problem6a} \\
    & \ \text{s.t.} \, (\mathrm{\ref{eq:Problem5c}}), (\mathrm{\ref{eq:Problem5d}}), \label{eq:Problem6b} \\
    & \ \quad \ \, \left( a_n f_{n, \mathrm{m}}^{-b_n} + c_n f_{n, \mathrm{c}}^{-d_n} \right) D_n^{-1} \leq 1, \forall n \in \mathcal{N}_0, \label{eq:Problem6c} \\
    & \ \quad \ \sum_{n \in \mathcal{N}_0} \big( \kappa_{n, \mathrm{m}} a_n f_{n, \mathrm{m}}^{3 - b_n} + \kappa_{n, \mathrm{c}} c_n f_{n, \mathrm{c}}^{3 - d_n} + \sigma_n a_n f_{n, \mathrm{m}}^{-b_n} + \nonumber \\
    & \sigma_n c_n f_{n, \mathrm{c}}^{-d_n} + \kappa_{n, \mathrm{m}} c_n f_{n, \mathrm{m}}^3 f_{n, \mathrm{c}}^{-d_n} + \kappa_{n, \mathrm{c}} a_n f_{n, \mathrm{c}}^3 f_{n, \mathrm{m}}^{-b_n} \big) y^{-1} \leq 1. \label{eq:Problem6d}
\end{alignat}
\end{subequations}
Notice that the objective function (\ref{eq:Problem6a}) is a monomial function, and the constraints (\ref{eq:Problem6b}), (\ref{eq:Problem6c}) and (\ref{eq:Problem6d}) give the upper bounds for posynomial functions. Therefore, problem (\ref{eq:Problem6}) is a GP problem, and the optimal memory frequency and computing frequency for problem (\ref{eq:Problem5}) can be efficiently obtained using an interior-point algorithm \cite{Boyd55}. The arithmetic cost of each step in solving problem (\ref{eq:Problem6}) is $\mathcal{O}\left(V_1\vert\mathcal{N}_0\vert^2\right)$, and the number of iterations for the interior-point algorithm is $\mathcal{O}\left(\sqrt{V_1}\log\left(1/\epsilon\right)\right)$, where $V_1 = 5\vert\mathcal{N}_0\vert + 1$ and $\epsilon$ represent the number of constraints and the convergence accuracy, respectively \cite{Nesterov56}. Therefore, the overall computational complexity of solving problem (\ref{eq:Problem5}) is $\mathcal{O}\left(\vert\mathcal{N}_0\vert^{3.5}\log\left(1/\epsilon\right)\right)$. 

\subsection{Uplink Bandwidth Allocation}
The edge subproblem is to minimize the total energy consumption of devices set to $x_n = 1$ with the deadline constraint by optimizing the uplink bandwidth. Specifically, the uplink bandwidth allocation subproblem is formulated as
\begin{subequations}\label{eq:Problem8}
\begin{alignat}{2}
    & E_{\mathrm{O}}^{\ast} \triangleq \min _{\mathbf{b}} \sum_{n \in \mathcal{N}_1} \frac{p_n s_n}{B_n \eta_n} \label{eq:Problem8a} \\
    & \ \text{s.t.} \, \frac{s_n}{B_n \eta_n} + t_n^{\mathrm{E}} \leq D_n, \forall n \in \mathcal{N}_1, \label{eq:Problem8b} \\
    & \ \quad \ \, \sum_{n \in \mathcal{N}_1} B_n \leq B, \label{eq:Problem8c} \\
    & \ \quad \ \, B_n \geq 0, \forall n \in \mathcal{N}_1. \label{eq:Problem8d}
\end{alignat}
\end{subequations}
Since the objective function and constraints of problem (\ref{eq:Problem8}) are all convex, problem (\ref{eq:Problem8}) is a standard convex optimization problem \cite{Nan13}, \cite{Li57}, and the optimal bandwidth allocation can be solved with an interior-point algorithm. The computational complexity of solving problem (\ref{eq:Problem8}) for each iteration is $\mathcal{O}\left( V_2 \vert\mathcal{N}_1\vert^2 \right)$, and the number of iterations for the IPM algorithm is $\mathcal{O}(\sqrt{V_2}\log(1/\epsilon))$, where $V_2 = 2\vert\mathcal{N}_1\vert + 1$ and $\epsilon$ represent the number of constraints and the convergence accuracy, respectively \cite{Nesterov56}. Therefore, the overall computational complexity of solving problem (\ref{eq:Problem8}) is $\mathcal{O}(\vert\mathcal{N}_1\vert^{3.5}\log(1/\epsilon))$. 

The overall procedure for solving problem (\ref{eq:Problem1}) with the given offloading policy is presented in Algorithm \ref{alg:1}. The optimal solution to problem (\ref{eq:Problem5}) is derived by solving problem (\ref{eq:Problem6}) with relatively low computational complexity, while the optimal bandwidth is derived by solving problem (\ref{eq:Problem8}). Therefore, Algorithm \ref{alg:1} can obtain the optimal memory frequency, computing frequency and bandwidth allocation under a given offloading policy in polynomial time complexity.
\begin{algorithm}[t]
    \renewcommand{\algorithmicrequire}{\textbf{Input:}}      
    \renewcommand{\algorithmicensure}{\textbf{Output:}}      
    \caption{Procedure for Solving Problem (\ref{eq:Problem1}) with the Given Offloading Policy}
    \label{alg:1}
    \begin{algorithmic}[1]
        \Require Number of devices $N$, offloading policy $\mathbf{x}$, task deadline $D_n$, transmission power $p_n$, edge inference time $t_n^{\mathrm{E}}$, memory frequency range $[f_{n, \min}, f_{n, \max}]$, computing frequency range $[F_{n, \min}, F_{n, \max}]$, total bandwidth $B$;
        \Ensure Optimal memory frequency $\mathbf{f_{M}^{\ast}} \triangleq \left\{ f_{n, \mathrm{m}}^{\ast} \right\}_{n \in \mathcal{N}}$, computing frequency $\mathbf{f_{C}^{\ast}} \triangleq \left\{ f_{n, \mathrm{c}}^{\ast} \right\}_{n \in \mathcal{N}}$, uplink bandwidth $\mathbf{b^{\ast}} \triangleq \left\{ B_n^{\ast} \right\}_{n \in \mathcal{N}}$, local energy consumption $E_{\mathrm{L}}^{\ast}$, and transmission energy consumption $E_{\mathrm{O}}^{\ast}$; 
        \State Local subproblem solving: for $n \in \mathcal{N}_0$, transform problem (\ref{eq:Problem5}) into an equivalent GP problem (\ref{eq:Problem6}), and obtain $f_{n, \mathrm{m}}^{\ast}$, ${f_{n, \mathrm{c}}^{\ast}}$ and $E_{\mathrm{L}}^{\ast}$ with the interior-point algorithm; 
        \State Edge subproblem solving: for $n \in \mathcal{N}_1$, derive $B_n^{\ast}$ and $E_{\mathrm{O}}^{\ast}$ by solving problem (\ref{eq:Problem8}) with the interior-point algorithm; 
    \end{algorithmic}
\end{algorithm}

\subsection{Task Offloading}
An efficient CD algorithm is used to optimize the offloading policy $\mathbf{x} = \left(x_1,\dots,x_N\right)$. The CD algorithm optimizes $\mathbf{x}$ along the descent direction of only one variable $x_i$ each time to find a local minimum \cite{Rao58}. Specifically, the CD algorithm first starts from a feasible offloading policy $\mathbf{x}^{\left(0\right)}$ randomly, where each element takes a value of either $0$ or $1$. Denote $\mathbf{x}^{\left(l-1\right)}$ as the binary offloading policy in the $\left(l-1\right)$th iteration, $l = 1, 2, \cdots$. Correspondingly, denote $V\left(\mathbf{x}^{\left(l-1\right)}\right)$ as the optimal value of problem (\ref{eq:Problem1}) given $\mathbf{x}^{\left(l-1\right)}$ based on Algorithm \ref{alg:1}. Let $R_j^{\left(l\right)}$ denote the reward of mobile device $j$ to exchange its current offloading policy in the $l$th iteration, which is defined as the reduction for the objective value of problem (\ref{eq:Problem1}) after the exchange, i.e.,
\begin{equation}
    R_j^{\left(l\right)} = V\left(\mathbf{x}^{\left(l-1\right)}\right) - V\left(\mathbf{x}^{\left(l-1\right)}\left(j\right)\right), \label{eq:CDReward}
\end{equation}
where $\mathbf{x}^{\left(l-1\right)}\left(j\right)$ denotes the offloading policy after mobile device $j$ exchanges its current policy, i.e.,
\begin{equation}
    \mathbf{x}^{\left(l-1\right)}\left(j\right) = \left(x_1^{\left(l-1\right)}, \cdots, x_j^{\left(l-1\right)} \oplus 1, \cdots, x_N^{\left(l-1\right)} \right). \label{eq:CDPolicy}
\end{equation}
Here, the operator $\oplus$ represents the modulo-2 addition, e.g., $1 \oplus 1 = 0$. Let the mobile device with the largest reward exchange its offloading policy if the reward is positive. Then, we obtain the offloading policy in the $l$th iteration $\mathbf{x}^{\left(l\right)} = \mathbf{x}^{\left(l-1\right)}\left(j_l^{\ast}\right)$ if $R_{j_l^{\ast}}^{\left(l\right)} > 0$, where $j_l^{\ast} = \argmax_{j \in \mathcal{N}} R_j^{\left(l\right)}$. The pseudo-code of the proposed CD algorithm is presented in Algorithm \ref{alg:2}, and the convergence analysis of Algorithm \ref{alg:2} is presented in the following proposition.
\begin{algorithm}[t]
\renewcommand{\algorithmicrequire}{\textbf{Input:}} 
\renewcommand{\algorithmicensure}{\textbf{Output:}} 
\caption{Coordinate Descent Algorithm for Offloading Policy Optimization}
\label{alg:2}
\begin{algorithmic}[1]
    \Require Initial offloading policy $\mathbf{x}^{\left(0\right)}$;
    \Ensure An approximate solution $\{\hat{\mathbf{x}}, \hat{\mathbf{f}}_{\mathrm{M}}, \hat{\mathbf{f}}_{\mathrm{C}}, \hat{\mathbf{b}}\}$ to problem (\ref{eq:Problem1});
    \State \textbf{Initialization:} $l \leftarrow 0$;
    \While{true}
        \State $l \leftarrow l + 1$;
        \For{each mobile device $j$}
            \State Calculate $R_j^{\left(l\right)}$ in (\ref{eq:CDReward}) using Algorithm \ref{alg:1};
        \EndFor
        \State $v_{l}^{\ast} \leftarrow \max_{j \in \mathcal{N}} R_j^{\left(l\right)}$;
        \State $j_l^{\ast} \leftarrow \argmax_{j \in \mathcal{N}} R_j^{\left(l\right)}$;
        \If{$v_l^{\ast} > 0$}
            \State Update $\mathbf{x}^{\left(l\right)} \leftarrow \mathbf{x}^{\left(l-1\right)}\left(j_l^{\ast}\right)$ using (\ref{eq:CDPolicy});
        \Else
            \State $\mathbf{x}^{\left(l\right)} \leftarrow \mathbf{x}^{\left(l-1\right)}$;
            \State \textbf{break};
        \EndIf
    \EndWhile
    \State Set $\hat{\mathbf{x}}\leftarrow \mathbf{x}^{(l)}$, and obtain the solution $\{\hat{\mathbf{f}}_{\mathrm{M}}, \hat{\mathbf{f}}_{\mathrm{C}}, \hat{\mathbf{b}}\}$ with the given $\hat{\mathbf{x}}$ using Algorithm \ref{alg:1};
    \State \Return An approximate solution $\{\hat{\mathbf{x}}, \hat{\mathbf{f}}_{\mathrm{M}}, \hat{\mathbf{f}}_{\mathrm{C}}, \hat{\mathbf{b}}\}$ to problem (\ref{eq:Problem1});
\end{algorithmic}
\end{algorithm}

\emph{Proposition 1:} Algorithm \ref{alg:2} converges in a finite number of iterations and returns a local optimum of problem (\ref{eq:Problem1}).

\begin{IEEEproof}
At the $l$th iteration, Algorithm \ref{alg:2} evaluates the objective reduction obtained by flipping the offloading policy of device $j$ based on (\ref{eq:CDReward}), where $\mathbf{x}^{\left(l-1\right)}\left(j\right)$ denotes the offloading policy obtained by flipping the $j$th element of $\mathbf{x}^{\left(l-1\right)}$. Let $v_l^{\ast} = \max_{j \in \mathcal{N}} R_j^{(l)}$. If $v_l^{\ast} > 0$, Algorithm \ref{alg:2} updates the offloading policy by flipping the coordinate with the largest reward. Therefore, the objective value satisfies
\begin{equation*}
    V\left(\mathbf{x}^{\left(l\right)}\right) = V\left(\mathbf{x}^{\left(l-1\right)}\right) - v_l^{\ast} < V\left(\mathbf{x}^{\left(l-1\right)}\right), 
\end{equation*}
which indicates that the objective value strictly decreases after each accepted update. Since the offloading policy is binary, the number of possible offloading policies is finite and no larger than $2^N$. Moreover, Algorithm \ref{alg:2} cannot revisit any previously visited offloading policy, because the objective value strictly decreases after each accepted update. Therefore, Algorithm \ref{alg:2} must terminate within a finite number of iterations. When Algorithm \ref{alg:2} terminates, $v_l^{\ast} \leq 0$, and flipping any single coordinate of the current offloading policy cannot further reduce the objective value. Therefore, the obtained offloading policy is a local optimum. Since Algorithm \ref{alg:1} provides the optimal resource allocation for the obtained offloading policy, Algorithm \ref{alg:2} returns a local optimum of problem (\ref{eq:Problem1}).
\end{IEEEproof}
    
Finally, we analyze the computational complexity of Algorithm \ref{alg:2}. Since each iteration of the CD algorithm needs to run Algorithm \ref{alg:1} $N$ times, the computational complexity of Algorithm \ref{alg:2} depends on Algorithm \ref{alg:1}. If Algorithm \ref{alg:2} requires $L$ iterations to converge, the overall computational complexity of Algorithm \ref{alg:2} is $\mathcal{O}(LN^{4.5}\log(1/\epsilon))$.

\section{Simulation Results}\label{sec:Simulation}
In this section, we first present the simulation setting. Then, we illustrate and analyze the performance of our proposed solution for local inference. Next, we empirically evaluate the convergence of Algorithm \ref{alg:2}, and give the performance evaluation under different parameter settings. Finally, we analyze the impact of deadline and bandwidth on the optimization of memory frequency and computing frequency in a specific case.

\subsection{Simulation Setting}
We consider an edge intelligence system similar to \cite{Nan31}, where the edge node is at the center of a $400 ~\mathrm{m} \times 400 ~\mathrm{m}$ square area, and all mobile devices are randomly placed within the area. The path loss (in dB) between mobile device $n$ and the edge node is characterized as $\mathrm{PL} = 38 + 30 \log_{10} r_n$ \cite{3GPP59}, where $r_n$ (in meters) is the distance between mobile device $n$ and the edge node. The transmission power is 1 W, and the noise power spectral density is $N_0 = -174$ dBm/Hz \cite{Nan31}, \cite{Geng60}. 

To evaluate the proposed algorithm, we conduct experiments on three hardware platforms with diverse DNN models. Specifically, ResNet152 \cite{He28}, VGG19 \cite{Simonyan29} and LSTM \cite{Hochreiter46} are deployed on Jetson TX1 and evaluated on the CIFAR-100 and UCI HAR datasets. ResNet152 and ViT-B/16 \cite{Dosovitskiy30} are deployed on Jetson Orin Nano for inference on the CIFAR-100 dataset. Swin-B \cite{Liu48} and GNN \cite{Scarselli61} are deployed on Jetson Orin NX and evaluated on the ImageNet and ogbn-arxiv datasets from the Open Graph Benchmark (OGB) \cite{Hu62}. Each VM is configured with an NVIDIA GeForce RTX 4080. The corresponding parameters and configurations are summarized in Table \ref{tab:II}.

\captionsetup[table]{labelsep=newline}
\begin{table*}[t]
    \scriptsize 
    \caption{\label{tab:II}\textsc{Configurations and Parameters of DNNs and Devices}}
	\centering
	\begin{tabular}{p{1.3cm}<{\centering} | p{1.8cm}<{\centering} | p{1.5cm}<{\centering} | p{1.6cm}<{\centering} | p{0.8cm}<{\centering} | p{0.8cm}<{\centering} | p{0.8cm}<{\centering} | p{0.8cm}<{\centering} | p{0.8cm}<{\centering} | p{0.8cm}<{\centering} | p{0.8cm}<{\centering} | p{0.8cm}<{\centering}}
		\hline
		\textcolor{black}{DNN Model} &\textcolor{black}{Mobile Device} &\textcolor{black}{$f_{n, \mathrm{m}}$} &\textcolor{black}{$f_{n, \mathrm{c}}$} &\textcolor{black}{$a_n$} &\textcolor{black}{$b_n$} &\textcolor{black}{$c_n$} &\textcolor{black}{$d_n$} &\textcolor{black}{$\kappa_{n, \mathrm{m}}$} &\textcolor{black}{$\kappa_{n, \mathrm{c}}$} &\textcolor{black}{$\sigma_n$} &\textcolor{black}{$t_n^{\mathrm{E}}$} \\ 
        \hline
		\textcolor{black}{ResNet152} &\multirow{3}{*}{\textcolor{black}{Jetson TX1}} &\multirow{3}{*}{\textcolor{black}{[0.1, 1.6] GHz}} &\multirow{3}{*}{\textcolor{black}{[0.1, 1.0] GHz}} &\textcolor{black}{0.039} &\textcolor{black}{1.552} &\textcolor{black}{0.211} &\textcolor{black}{0.358} &\multirow{3}{*}{\textcolor{black}{0.152}} &\multirow{3}{*}{\textcolor{black}{1.115}} &\multirow{3}{*}{\textcolor{black}{2.097}} &\textcolor{black}{12 ms} \\
        \cline{1-1}\cline{5-8}\cline{12-12}
        \textcolor{black}{VGG19} &  &  &  &\textcolor{black}{0.110} &\textcolor{black}{1.296} &\textcolor{black}{0.099} &\textcolor{black}{0.629} &  &  &  &\textcolor{black}{4 ms} \\
        \cline{1-1}\cline{5-8}\cline{12-12}
        \textcolor{black}{LSTM} &  &  &  &\textcolor{black}{0.051} &\textcolor{black}{1.193} &\textcolor{black}{0.029} &\textcolor{black}{0.370} &  &  &  &\textcolor{black}{1 ms} \\
        \hline
        \textcolor{black}{ResNet152} &\multirow{2}{*}{\textcolor{black}{Jetson Orin Nano}} &\multirow{2}{*}{\textcolor{black}{[0.2, 2.1] GHz}} &\multirow{2}{*}{\textcolor{black}{[0.3, 0.63] GHz}} &\textcolor{black}{0.012} &\textcolor{black}{1.426} &\textcolor{black}{0.112} &\textcolor{black}{0.036} &\multirow{2}{*}{\textcolor{black}{0.033}} &\multirow{2}{*}{\textcolor{black}{1.268}} &\multirow{2}{*}{\textcolor{black}{1.936}} &\textcolor{black}{12 ms} \\
        \cline{1-1}\cline{5-8}\cline{12-12}
		\textcolor{black}{ViT-B/16} &  &  &  &\textcolor{black}{0.012} &\textcolor{black}{1.638} &\textcolor{black}{0.043} &\textcolor{black}{0.241} &  &  &  &\textcolor{black}{6 ms} \\
        \hline
        \textcolor{black}{GNN} &\multirow{2}{*}{\textcolor{black}{Jetson Orin NX}} &\multirow{2}{*}{\textcolor{black}{[0.2, 3.2] GHz}} &\multirow{2}{*}{\textcolor{black}{[0.3, 1.2] GHz}} &\textcolor{black}{0.060} &\textcolor{black}{1.423} &\textcolor{black}{0.081} &\textcolor{black}{0.700} &\multirow{2}{*}{\textcolor{black}{0.043}} &\multirow{2}{*}{\textcolor{black}{0.698}} &\multirow{2}{*}{\textcolor{black}{4.250}} &\textcolor{black}{5 ms} \\
        \cline{1-1}\cline{5-8}\cline{12-12}
        \textcolor{black}{Swin-B} &  &  &  &\textcolor{black}{0.020} &\textcolor{black}{1.445} &\textcolor{black}{0.131} &\textcolor{black}{0.159} &  &  &  &\textcolor{black}{26 ms} \\
        \hline
	\end{tabular}
    \vspace{-0.2cm}
\end{table*}
\begin{figure*}[t]
    \captionsetup{font={small}}
    \centering
    \begin{subfigure}{0.32\linewidth}
        \centering
        \includegraphics[width=\linewidth]{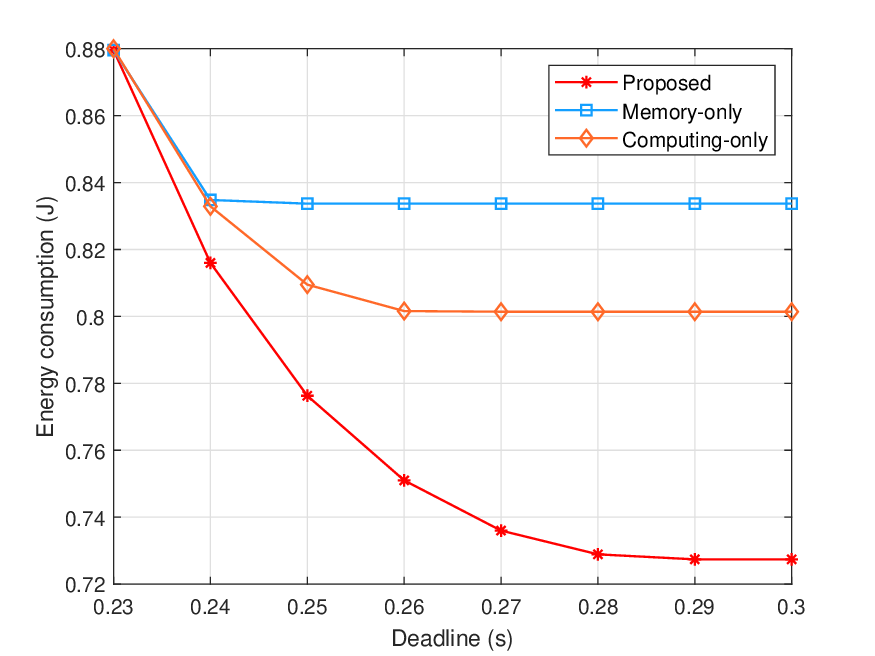}
        \caption{ResNet152 on Jetson TX1}
        \label{fig:E_ResNet152TX1}
    \end{subfigure}
    \hfill
    \begin{subfigure}{0.32\linewidth}
        \centering
        \includegraphics[width=\linewidth]{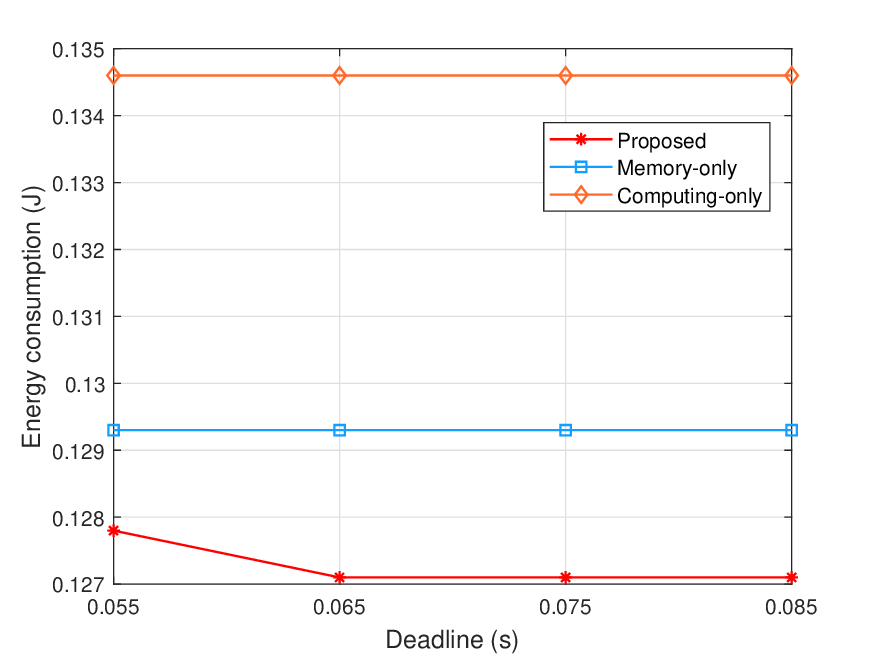}
        \caption{ViT-B/16 on Jetson Orin Nano}
        \label{fig:E_ViTOrin}
    \end{subfigure}
    \hfill
    \begin{subfigure}{0.32\linewidth}
        \centering
        \includegraphics[width=\linewidth]{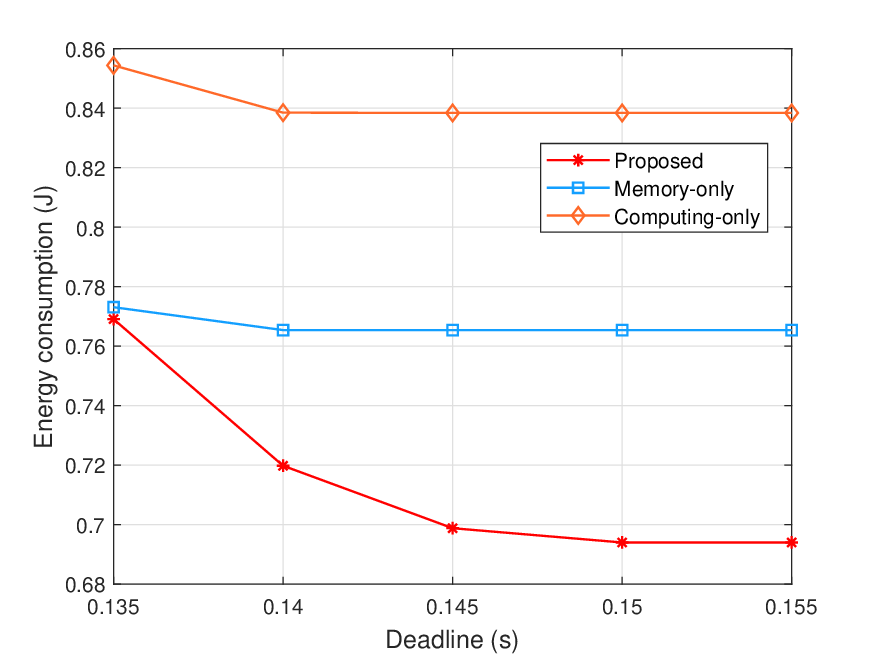}
        \caption{Swin-B on Jetson Orin NX}
        \label{fig:E_SwinBNX}
    \end{subfigure}
    \caption{Comparison for the energy consumption of local inference with different deadlines for ResNet152 on Jetson TX1, ViT-B/16 on Jetson Orin Nano and Swin-B on Jetson Orin NX.}\label{fig:Local_Compare_Energy}
    \vspace{-0.3cm}
\end{figure*}

\begin{figure*}[t]
    \captionsetup{font={small}}
    \centering
    \begin{subfigure}{0.32\linewidth}
        \centering
        \includegraphics[width=\linewidth]{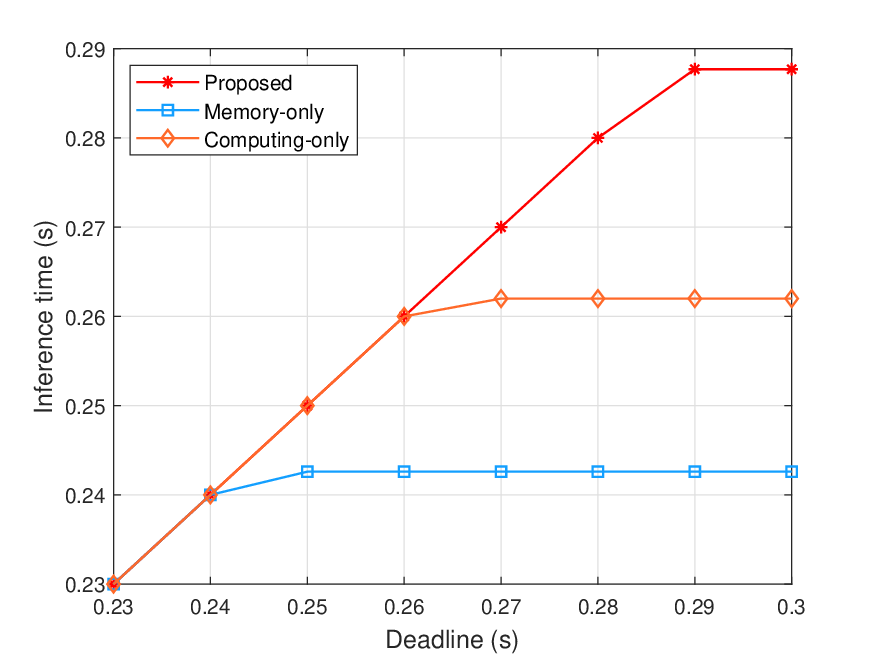}
        \caption{ResNet152 on Jetson TX1}
        \label{fig:T_ResNet152TX1}
    \end{subfigure}
    \hfill
    \begin{subfigure}{0.32\linewidth}
        \centering
        \includegraphics[width=\linewidth]{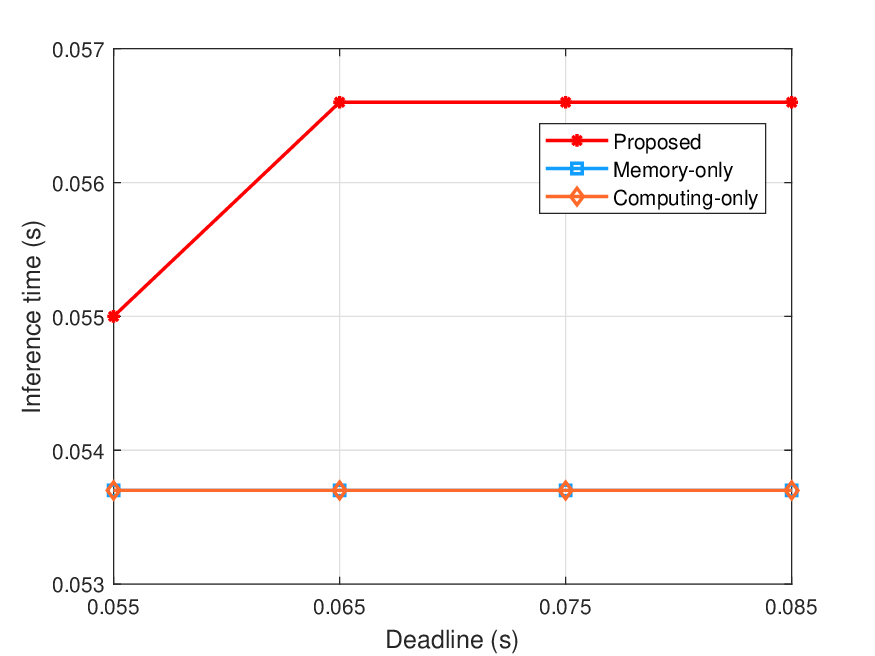}
        \caption{ViT-B/16 on Jetson Orin Nano}
        \label{fig:T_ViTOrin}
    \end{subfigure}
    \hfill
    \begin{subfigure}{0.32\linewidth}
        \centering
        \includegraphics[width=\linewidth]{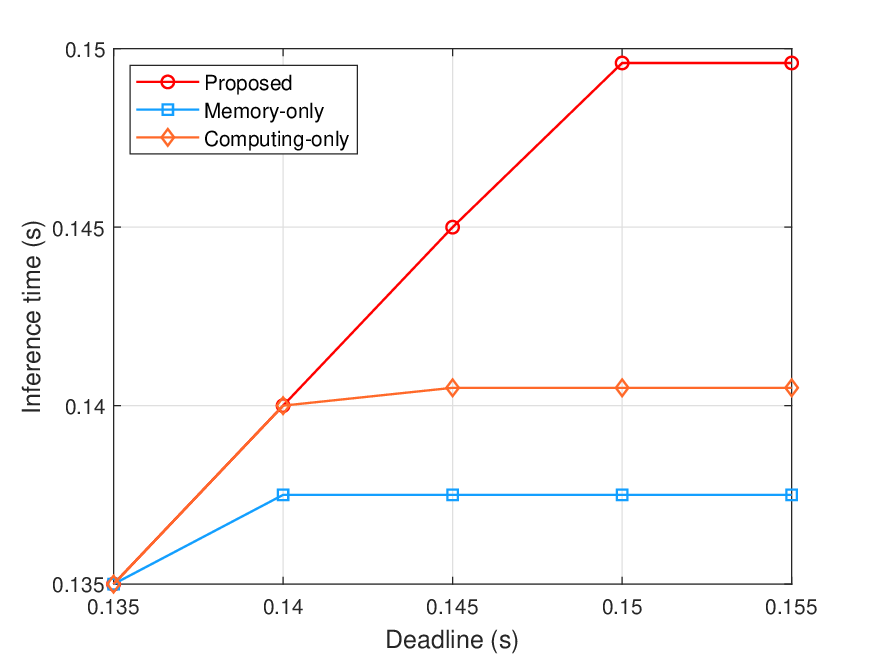}
        \caption{Swin-B on Jetson Orin NX}
        \label{fig:T_SwinBNX}
    \end{subfigure}
    \caption{Comparison for the local inference time with different deadlines for ResNet152 on Jetson TX1, ViT-B/16 on Jetson Orin Nano and Swin-B on Jetson Orin NX.}\label{fig:Local_Compare_InferenceTime}
    \vspace{-0.3cm}
\end{figure*}

\subsection{Comparison for Local Inference}
We first compare the performance of our proposed solution with two benchmark methods under different deadlines for local inference. Specifically, the memory-only solution is given as a benchmark, which fixes the computing frequency to the maximum value and only optimizes the memory frequency. The computing-only solution is provided as another benchmark, which fixes the memory frequency to the maximum value and only optimizes the computing frequency.

Fig. \ref{fig:Local_Compare_Energy} presents the energy consumption of one mobile device executing local inference for our proposed solution, memory-only solution and computing-only solution. Based on the analysis in Section \ref{sec:Solution2}, our proposed solution is optimal, which can be derived by solving the equivalent GP problem and performs better than the other two benchmark solutions for the energy consumption. In Fig. \ref{fig:E_ResNet152TX1} and Fig. \ref{fig:E_SwinBNX}, the performance gap is relatively small with a small deadline. As the deadline increases, our proposed solution can further reduce the energy consumption, while the reductions achieved by the memory-only and computing-only solutions are limited. Specifically, compared with the memory-only and computing-only solutions, our proposed solution can reduce the energy consumption by $12.6\%$ and $9.0\%$ with $D_n = 280$ ms for ResNet152 on Jetson TX1, respectively. Compared with the memory-only and computing-only solutions, the proposed solution reduces the energy consumption by $9.3\%$ and $17.2\%$ with $D_n = 150$ ms for Swin-B on Jetson Orin NX, respectively. However, in Fig. \ref{fig:E_ViTOrin}, the energy consumption of our proposed solution does not significantly decrease with the deadline, where the corresponding performance gaps from the memory-only and computing-only solutions are only $1.7\%$ and $5.6\%$ with $D_n = 65$ ms for ViT-B/16 on Jetson Orin Nano, respectively. For the memory-only and computing-only solutions with larger deadlines, the energy consumption remains constant because the optimization of memory/computing frequency is limited when the computing/memory frequency is set to the maximum. In conclusion, by jointly optimizing the memory frequency and computing frequency, the energy consumption of local inference can be further reduced. 
\begin{figure*}[t]
    \captionsetup{font={small}}
    \centering
    \begin{subfigure}{0.32\linewidth}
        \centering
        \includegraphics[width=\linewidth]{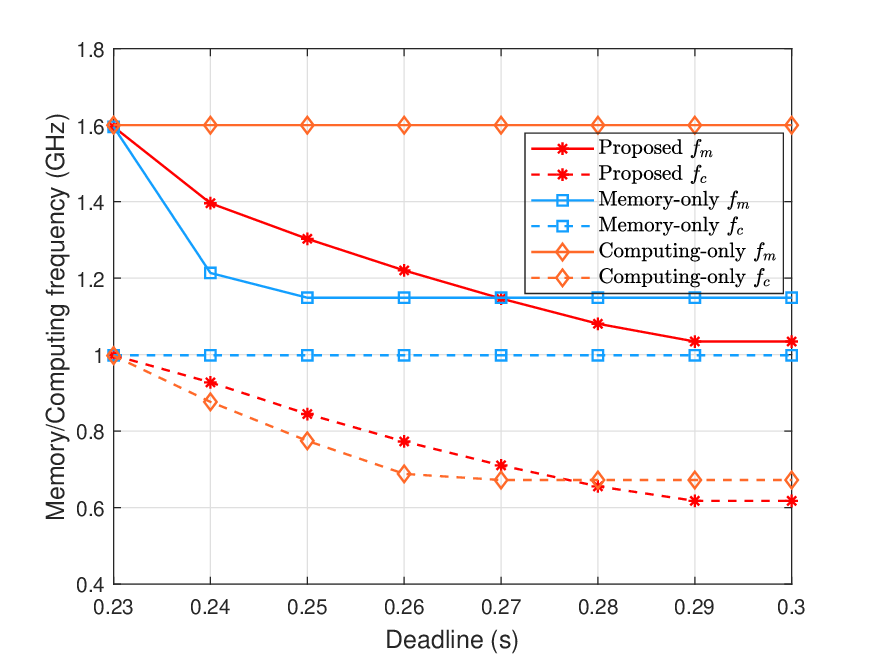}
        \caption{ResNet152 on Jetson TX1}
        \label{fig:F_ResNet152TX1}
    \end{subfigure}
    \hfill
    \begin{subfigure}{0.32\linewidth}
        \centering
        \includegraphics[width=\linewidth]{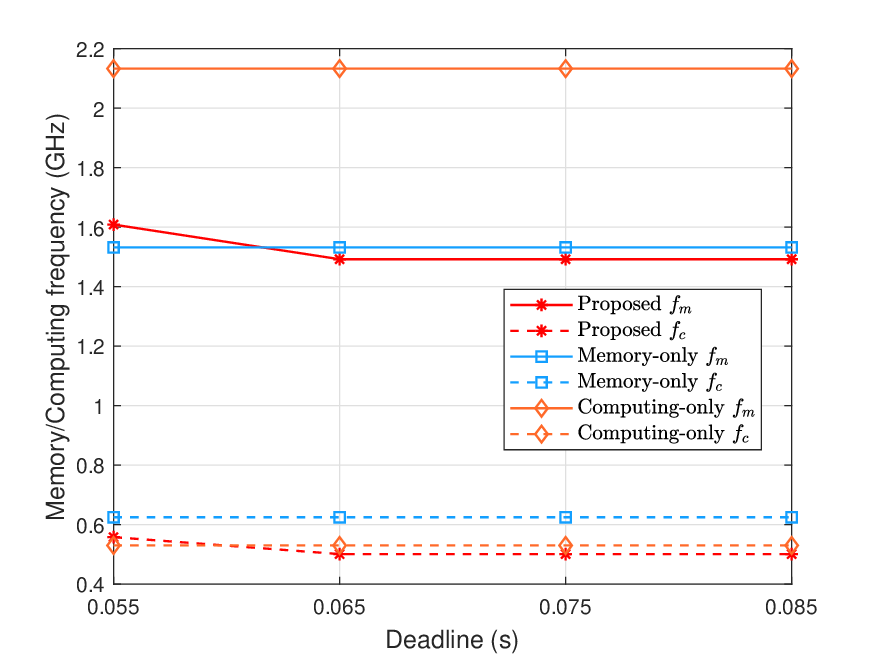}
        \caption{ViT-B/16 on Jetson Orin Nano}
        \label{fig:F_ViTOrin}
    \end{subfigure}
    \hfill
    \begin{subfigure}{0.32\linewidth}
        \centering
        \includegraphics[width=\linewidth]{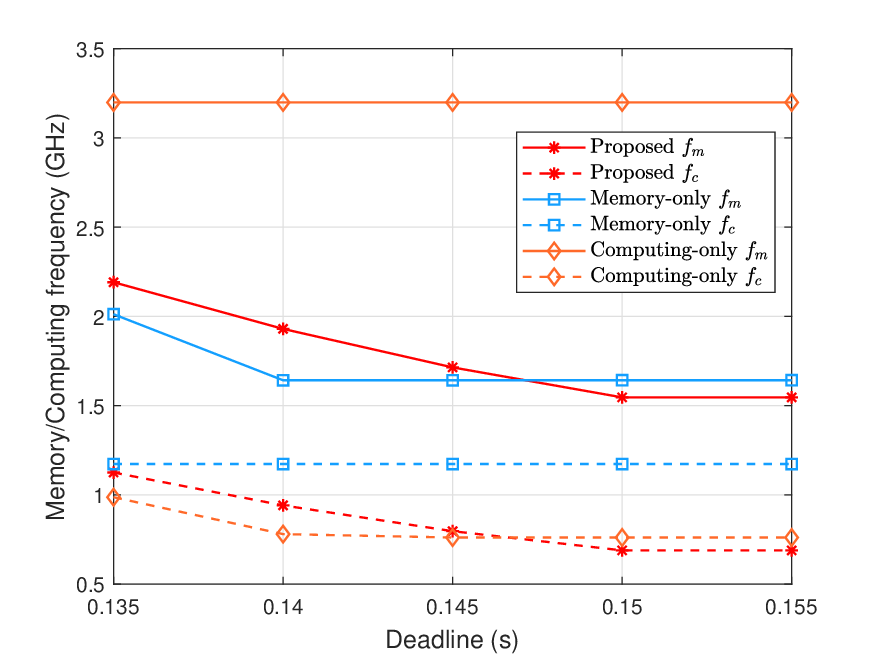}
        \caption{Swin-B on Jetson Orin NX}
        \label{fig:F_SwinBNX}
    \end{subfigure}
    \caption{Comparison for the memory frequency and computing frequency of local inference with different deadlines for ResNet152 on Jetson TX1, ViT-B/16 on Jetson Orin Nano and Swin-B on Jetson Orin NX.}\label{fig:Local_Compare_Frequency}
    \vspace{-0.3cm}
\end{figure*}

Fig. \ref{fig:Local_Compare_InferenceTime} illustrates the actual inference time of three solutions under different deadlines. In Fig. \ref{fig:T_ResNet152TX1} and Fig. \ref{fig:T_SwinBNX}, although the inference time of our proposed solution is equal to that of the computing-only solution with a relatively small deadline, their energy consumption differs due to distinct choices of memory frequency and computing frequency. As the deadline increases, the inference time of our proposed solution becomes larger than that of the memory-only or computing-only solution, indicating that our proposed solution fully exploits the available deadline to reduce the energy consumption of local inference. Moreover, the inference time of our proposed solution remains constant when the deadline further increases, which is attributed to the complicated and non-monotonic objective function in problem (\ref{eq:Problem5}). In addition, the difference among the three solutions in Fig. \ref{fig:T_ViTOrin} is relatively small, which is consistent with the results shown in Fig. \ref{fig:E_ViTOrin}. 

To further analyze the difference among the three solutions, we evaluate the specific values of memory frequency and computing frequency under different deadline constraints, as shown in Fig. \ref{fig:Local_Compare_Frequency}. As the deadline increases, the memory frequency and computing frequency derived by our proposed solution can be reduced. With a larger deadline, it is observed that our proposed solution obtains smaller memory frequencies and computing frequencies than the other two benchmark methods, leading to smaller energy consumption. In Fig. \ref{fig:F_ResNet152TX1} and Fig. \ref{fig:F_SwinBNX}, the frequency gap is much larger, which is consistent with the performance gain shown in Fig. \ref{fig:E_ResNet152TX1} and Fig. \ref{fig:E_SwinBNX}. The computing frequency gap among the three solutions is not significantly large in Fig. \ref{fig:F_ViTOrin}, since the parameters $c_n$ and $d_n$ for Jetson Orin Nano are relatively small. In summary, the memory frequency and computing frequency should be properly optimized to balance the inference time and energy consumption. Although the benchmark methods reduce the inference time by increasing the computing frequency and memory frequency, this latency reduction is not sufficient to compensate for the increased power consumption, resulting in higher overall energy consumption. Our proposed solution can provide a more appropriate selection for the memory frequency and computing frequency to reduce the energy consumption of local inference.

\subsection{Convergence of Algorithm \ref{alg:2}}
\begin{figure}[t]
  \captionsetup{font={small}}
  \centering
  \includegraphics[width=0.4\textwidth]{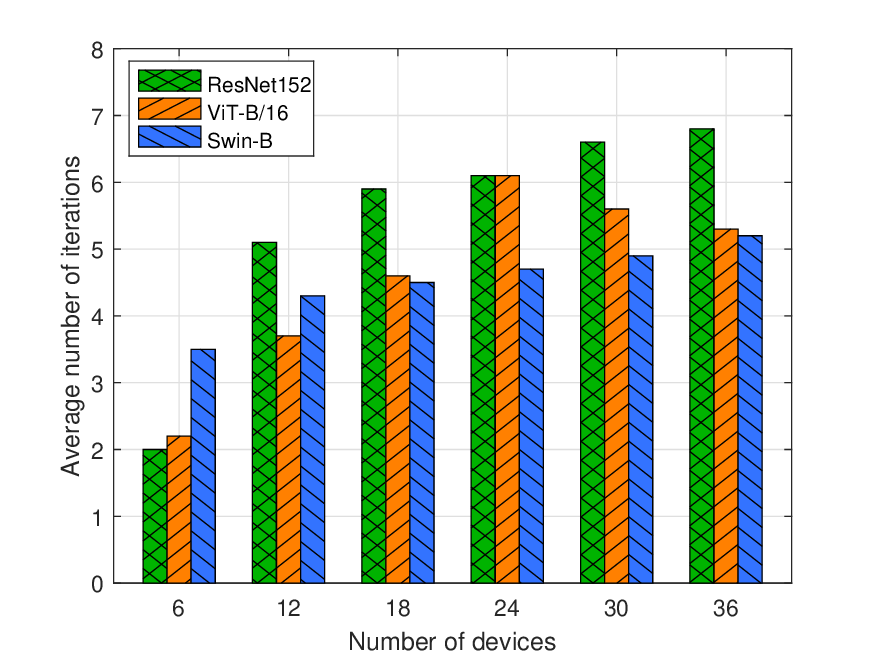}\\
  \caption{The average iterations of Algorithm \ref{alg:2} under different numbers of mobile devices for ResNet152, ViT-B/16 and Swin-B.}
  \label{fig:IterationAlg2}
  \vspace{-0.3cm}
\end{figure}

Fig. \ref{fig:IterationAlg2} presents the average number of iterations for Algorithm \ref{alg:2} under various numbers of devices. Although the number of iterations required for convergence is difficult to characterize analytically, the proposed CD-based algorithm empirically converges within a few iterations, even when the network scale increases to $N = 36$. Moreover, the average numbers of iterations required for the three DNN models are generally comparable. In addition, the average number of iterations required by Algorithm \ref{alg:2} increases marginally with the number of devices, which demonstrates the scalability and efficiency of the proposed algorithm. 

\subsection{Performance Evaluation of Algorithm \ref{alg:2}}
In this subsection, we evaluate the energy consumption performance of Algorithm \ref{alg:2} under various parameter settings. Specifically, we consider the following five benchmark policies for performance comparison:
\begin{itemize} 
    \item Optimal policy: the optimal memory frequency, computing frequency and bandwidth are obtained by Algorithm \ref{alg:1}, and the optimal offloading policy is determined by exhaustive search.
    \item Computing-only policy \cite{Nan13}: only the computing frequency is optimized in the local subproblem, and the CD algorithm is used to determine the offloading policy. 
    \item No-scaling policy: the memory frequency and computing frequency are set to the maximum values and not optimized in the local subproblem, and the CD algorithm is used to determine the offloading policy.
    \item DRL-based policy \cite{Zhang25}: the DQN is adopted to optimize the memory frequency, computing frequency and offloading policy, and the bandwidth allocation is obtained by the convex optimization method.
    \item Random policy: the optimal memory frequency, computing frequency and bandwidth are obtained by Algorithm \ref{alg:1}, and the offloading policy is given randomly.
\end{itemize}

\subsubsection{Impact of Device Numbers}
In Fig. \ref{fig:Energy_Device_TX}, Fig. \ref{fig:Energy_Device_Orin} and Fig. \ref{fig:Energy_Device_SwinB}, we compare the performance of different policies under various numbers of devices. It is observed that the average energy consumption increases with the number of devices when the deadline and communication resources are given. As expected, the average energy consumption of the optimal policy is the lowest, which provides a lower bound for other policies. When the number of devices is small (i.e., $N \leq 8$), our proposed algorithm can almost achieve the optimal performance. For example, when the number of devices $N = 8$, our proposed algorithm achieves the optimal energy consumption in Fig. \ref{fig:Energy_Device_TX}, while the performance gap between the optimal policy and our proposed algorithm is only $1.4\%$ in Fig. \ref{fig:Energy_Device_Orin}. In this case, the communication resources are sufficient and almost all mobile devices can offload their tasks to the edge server. As the number of devices increases, the performance of our proposed algorithm is still close to the optimal policy, and significantly outperforms other benchmark policies. This is because the communication resources are limited to support task offloading for all mobile devices, and some devices must execute local inference to meet the deadline constraint, leading to the increase of average energy consumption and performance gap. 

Specifically, for ResNet152, ViT-B/16 and Swin-B, the performance gap between our proposed algorithm and the optimal policy is no more than $6.4\%$, $1.5\%$ and $9.3\%$, respectively. Compared with the computing-only policy, our proposed algorithm can reduce the average energy consumption by up to $8.0\%$, $4.8\%$ and $15.6\%$, respectively, which verifies the benefit of jointly optimizing memory frequency and computing frequency. The average energy consumption of no-scaling policy is larger than that of computing-only policy, since the no-scaling policy sets both memory frequency and computing frequency to the maximum values during local inference. Compared with the DRL-based policy, the proposed algorithm achieves lower energy consumption with average reductions of $8.2\%$, $10.2\%$ and $16.0\%$, respectively. This is because the DRL-based policy learns the offloading and frequency decisions from sampled interactions and does not exploit the analytical structure of the problem. In fact, the learned policy is not stable and may violate the given deadline constraint. The random policy performs the worst in most cases because it randomly determines the offloading decision and thus cannot effectively exploit the available communication resources. 

\subsubsection{Impact of Deadline}
\begin{figure*}[!htb]
    \captionsetup{font={small}}
    \centering
    \begin{subfigure}{0.32\linewidth}
        \centering
        \includegraphics[width=\linewidth]{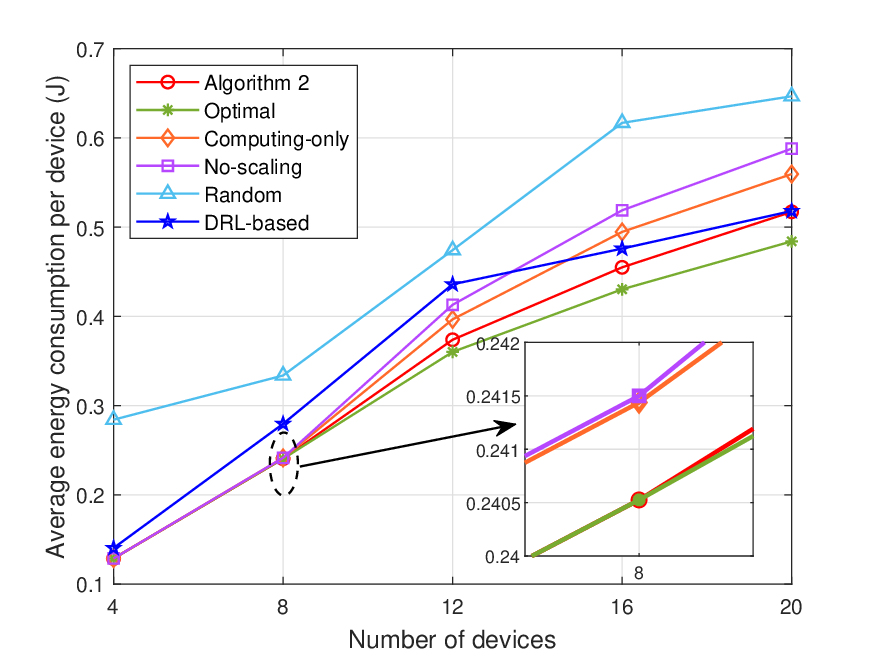}
        \caption{}
        \label{fig:Energy_Device_TX}
    \end{subfigure}
    \hfill
    \begin{subfigure}{0.32\linewidth}
        \centering
        \includegraphics[width=\linewidth]{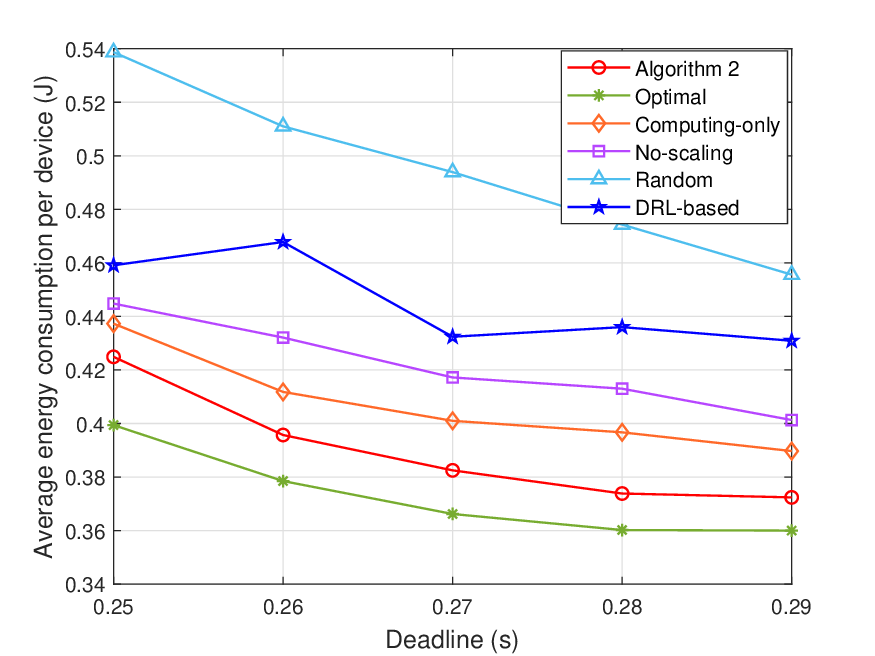}
        \caption{}
        \label{fig:Energy_Deadline_TX}
    \end{subfigure}
    \hfill
    \begin{subfigure}{0.32\linewidth}
        \centering
        \includegraphics[width=\linewidth]{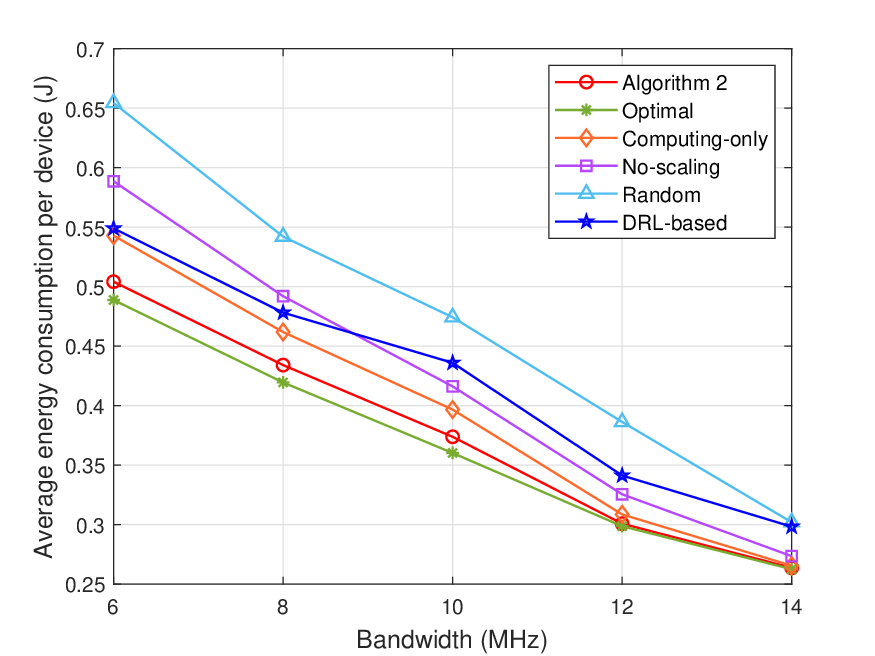}
        \caption{}
        \label{fig:Energy_Bandwidth_TX}
    \end{subfigure}
    \caption{Performance of the proposed algorithm for ResNet152 on Jetson TX1. (a) Average energy consumption per device under various numbers of devices with $D_n = 0.28$ s and $B = 10 $ MHz. (b) Average energy consumption per device under various deadlines with $N = 12$ and $B = 10$ MHz. (c) Average energy consumption per device under various bandwidth resources with $N = 12$ and $D_n = 0.28$ s.}
    \label{fig:Energy_MultiDevice_TX1}
    \vspace{-0.3cm}
\end{figure*}

\begin{figure*}[!htb]
	\captionsetup{font={small}}
	\centering
    \begin{subfigure}{0.32\linewidth}
        \centering
        \includegraphics[width=\linewidth]{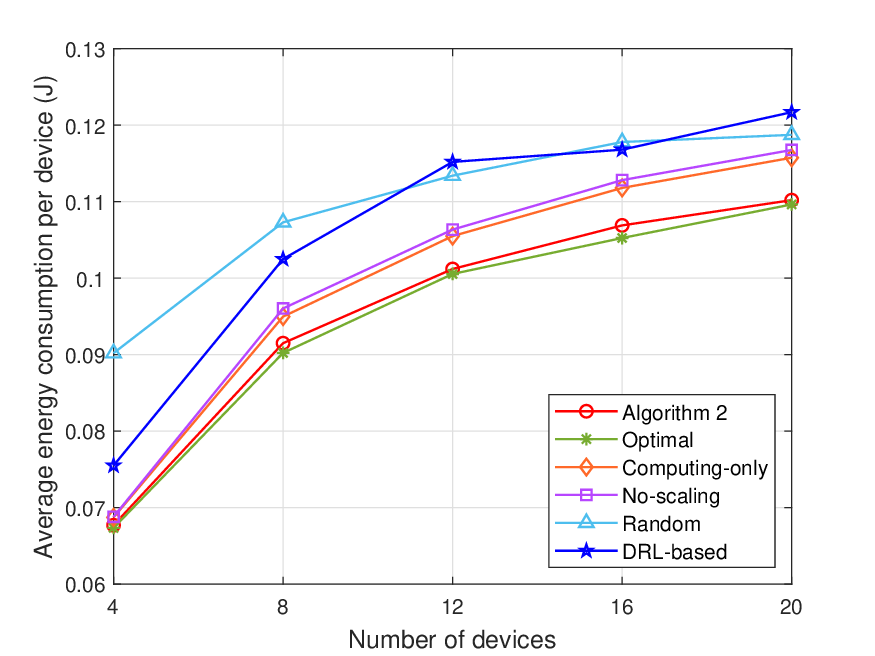}
        \caption{}
        \label{fig:Energy_Device_Orin}
    \end{subfigure}
    \hfill
    \begin{subfigure}{0.32\linewidth}
        \centering
        \includegraphics[width=\linewidth]{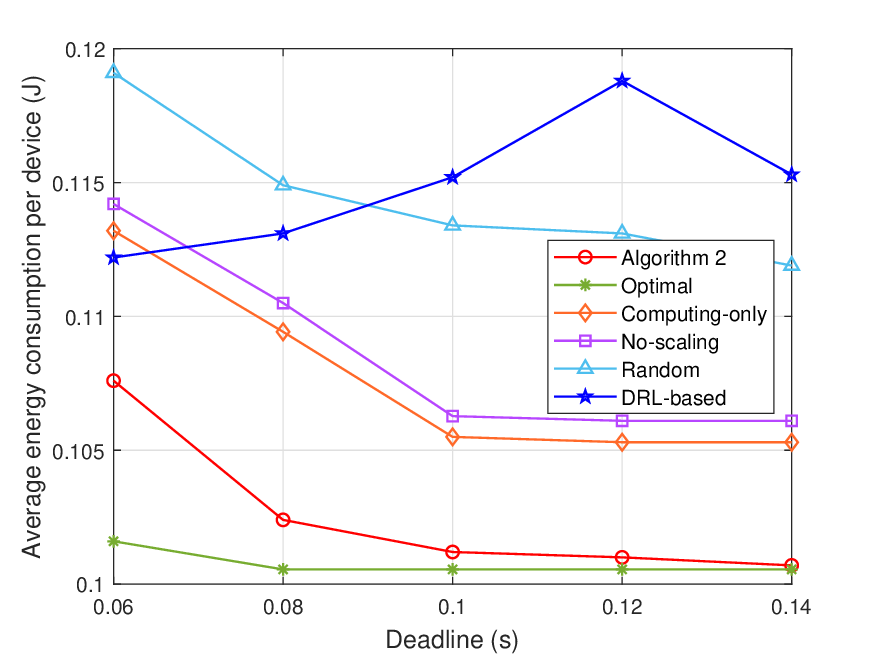}
        \caption{}
        \label{fig:Energy_Deadline_Orin}
    \end{subfigure}
    \hfill
    \begin{subfigure}{0.32\linewidth}
        \centering
        \includegraphics[width=\linewidth]{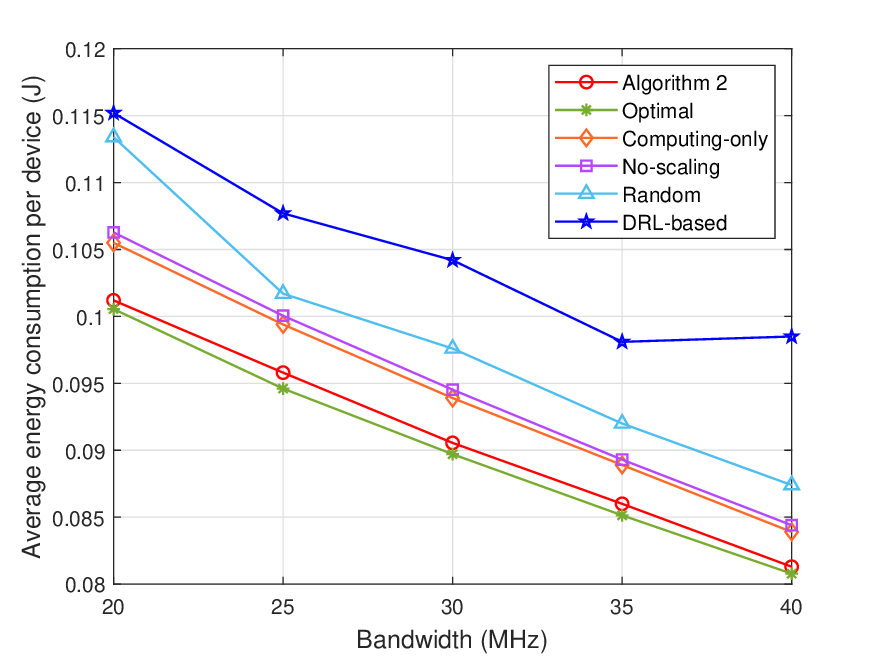}
        \caption{}
        \label{fig:Energy_Bandwidth_Orin}
    \end{subfigure}
	\caption{Performance of the proposed algorithm for ViT-B/16 on Jetson Orin Nano. (a) Average energy consumption per device under various numbers of devices with $D_n = 0.1$ s and $B = 20$ MHz. (b) Average energy consumption per device under various deadlines with $N = 12$ and $B = 20$ MHz. (c) Average energy consumption per device under various bandwidth resources with $N = 12$ and $D_n = 0.1$ s.}
	\label{fig:Energy_MultiDevice_Orin}
	\vspace{-0.3cm}
\end{figure*}

\begin{figure*}[!htb]
	\captionsetup{font={small}}
	\centering
    \begin{subfigure}{0.32\linewidth}
        \centering
        \includegraphics[width=\linewidth]{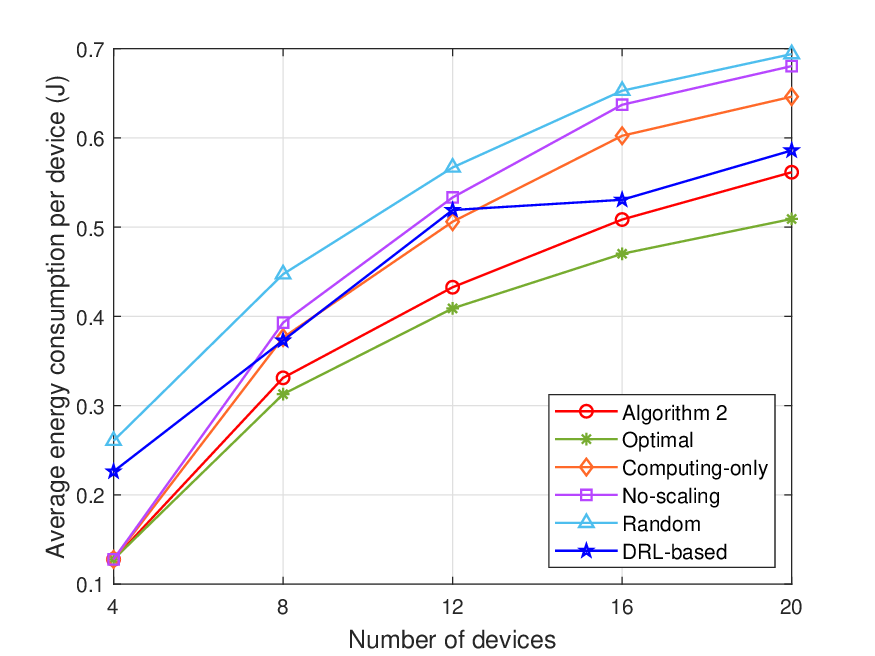}
        \caption{}
        \label{fig:Energy_Device_SwinB}
    \end{subfigure}
    \hfill
    \begin{subfigure}{0.32\linewidth}
        \centering
        \includegraphics[width=\linewidth]{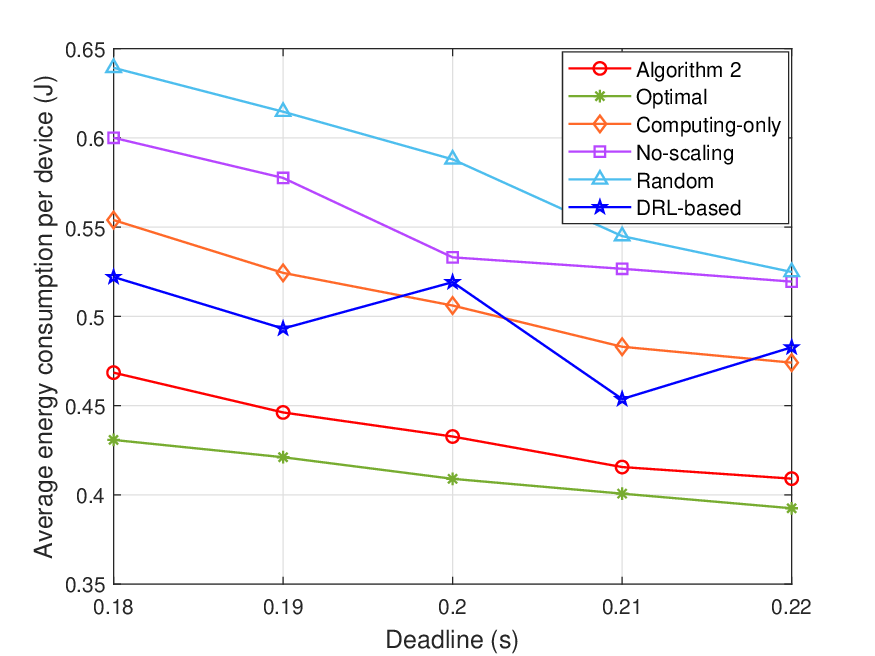}
        \caption{}
        \label{fig:Energy_Deadline_SwinB}
    \end{subfigure}
    \hfill
    \begin{subfigure}{0.32\linewidth}
        \centering
        \includegraphics[width=\linewidth]{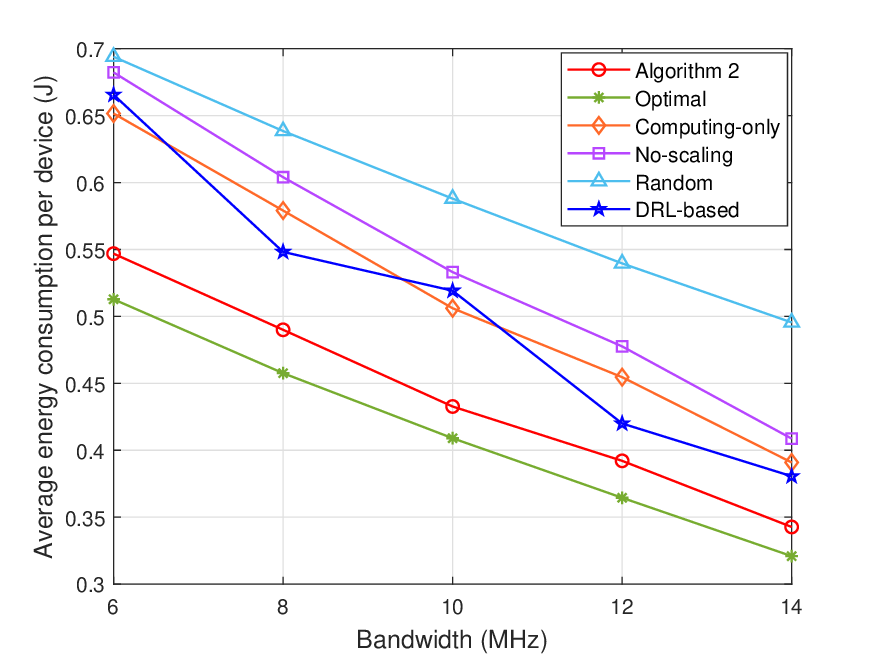}
        \caption{}
        \label{fig:Energy_Bandwidth_SwinB}
    \end{subfigure}
	\caption{Performance of the proposed algorithm for Swin-B on Jetson Orin NX. (a) Average energy consumption per device under various numbers of devices with $D_n = 0.2$ s and $B = 10$ MHz. (b) Average energy consumption per device under various deadlines with $N = 12$ and $B = 10$ MHz. (c) Average energy consumption per device under various bandwidth resources with $N = 12$ and $D_n = 0.2$ s.}
	\label{fig:Energy_MultiDevice_SwinB}
	\vspace{-0.3cm}
\end{figure*}

In Fig. \ref{fig:Local_Compare_Energy}, we have observed the impact of deadline on the energy consumption of local inference. To further investigate the impact of deadline on the average energy consumption per device in the multi-device case, we compare the performance of five policies under different deadlines, as shown in Fig. \ref{fig:Energy_Deadline_TX}, Fig. \ref{fig:Energy_Deadline_Orin} and Fig. \ref{fig:Energy_Deadline_SwinB}. As the deadline increases, the average energy consumption shows a decreasing trend. Increasing the deadline can provide more opportunities for devices to offload their inference tasks and reduce the energy consumption. The performance gap between our proposed algorithm and the optimal policy is no more than $6.0\%$, $5.6\%$ and $8.0\%$ for ResNet152, ViT-B/16 and Swin-B, respectively. Moreover, our proposed algorithm can achieve near-optimal performance when the deadline is large enough, since more devices offload their tasks and the difference only arises from the local inference. Compared with the computing-only policy, our proposed algorithm can reduce the average energy consumption by at most $5.8\%$, $6.4\%$ and $15.4\%$, respectively. Compared with the DRL-based policy, the proposed algorithm achieves average energy reductions of $12.5\%$, $10.7\%$ and $12.0\%$, respectively. The performance of the DRL-based policy is not stable because the discrete frequency and offloading decisions are adopted. In addition, there is a larger decrease for the average energy consumption in Fig. \ref{fig:Energy_Deadline_TX} and Fig. \ref{fig:Energy_Deadline_SwinB}, which is consistent with the results shown in Fig. \ref{fig:Local_Compare_Energy}. 

\subsubsection{Impact of Bandwidth}
In Fig. \ref{fig:Energy_Bandwidth_TX}, Fig. \ref{fig:Energy_Bandwidth_Orin} and Fig. \ref{fig:Energy_Bandwidth_SwinB}, the impact of bandwidth on average energy consumption for different policies is presented. Similar to the impact of deadline, increasing the bandwidth can significantly reduce the average energy consumption. This is because with more bandwidth resources, there are more devices offloading tasks to the edge node, and the energy consumption of communication is much smaller than that of executing local inference. For ResNet152, ViT-B/16 and Swin-B, the maximum performance gap between our proposed algorithm and the optimal policy is $3.6\%$, $1.2\%$ and $7.0\%$, respectively. In addition, our proposed algorithm outperforms the computing-only policy by up to $7.1\%$, $4.1\%$ and $16.1\%$, respectively, achieved under relatively small bandwidth resources (i.e., $B = 6$ MHz in Fig. \ref{fig:Energy_Bandwidth_TX}, $B = 20$ MHz in Fig. \ref{fig:Energy_Bandwidth_Orin} and $B = 6$ MHz in Fig. \ref{fig:Energy_Bandwidth_SwinB}). Another observation is that both our proposed algorithm and the computing-only policy achieve near-optimal performance when the bandwidth resources are abundant (e.g., $B = 14$ MHz in Fig. \ref{fig:Energy_Bandwidth_TX}), since almost all devices select to offload inference tasks in this case. Compared with the DRL-based policy, the proposed algorithm can reduce the energy consumption by $11.0\%$, $13.2\%$ and $12.3\%$ on average, respectively. In conclusion, our proposed algorithm can significantly reduce the average energy consumption by jointly optimizing the memory frequency, computing frequency and the allocation of bandwidth resources. 

\subsubsection{Ablation Studies}
\captionsetup[table]{labelsep=newline}
\begin{table*}[t]
    \scriptsize
    \centering
    \caption{\textsc{Average energy consumption (in J) of baselines with different settings for LSTM on Jetson TX1}}
	\setlength{\tabcolsep}{8pt}
	\renewcommand{\arraystretch}{1.1}
	\begin{tabular}{c|cccc|cccc|cccc}
		\hline
		\multirow{2}*{{Policy}} & \multicolumn{4}{c|}{{$N$ ($D = 0.1$ s, $B = 20$ MHz)}} & \multicolumn{4}{c|}{{$D$ ($N = 12$, $B = 20$ MHz)}} & \multicolumn{4}{c}{{$B$ ($N = 12$, $D = 0.1$ s)}} \\
		\multirow{2}*{} & 12 & 16 & 20 & 24 & 0.06 s & 0.08 s & 0.10 s & 0.12 s & 10 MHz & 15 MHz & 20 MHz & 25 MHz \\
		\hline
		Computing-only & 0.1398 & 0.1519 & 0.1595 & 0.1635 & \textbf{0.1574} & 0.1474 & 0.1398 & 0.1379 & 0.1624 & 0.1511 & 0.1398 & 0.1290          \\
		
		Memory-only & 0.1588 & 0.1754 & 0.1845 & 0.1892 & 0.1708 & 0.1684 & 0.1588 & 0.1544 & 0.1867 & 0.1726 & 0.1588 & 0.1433            \\
		
		Discrete-points & 0.1398 & 0.1520 & 0.1595 & 0.1636 & 0.1575 & 0.1474 & 0.1398 & 0.1380 & 0.1625 & 0.1512 & 0.1398 & 0.1292          \\
		\hline
		\textbf{Proposed (ours)} & \textbf{0.1395} & \textbf{0.1516} & \textbf{0.1591} & \textbf{0.1628} & \textbf{0.1574} & \textbf{0.1446} & \textbf{0.1395} & \textbf{0.1377} & \textbf{0.1619} & \textbf{0.1508} & \textbf{0.1395} & \textbf{0.1284}          \\
		\hline
	\end{tabular}
    \label{tab:IV}
\end{table*}

\begin{table*}[!htbp]
    \scriptsize
    \centering
	\caption{\textsc{Average energy consumption (in J) of baselines with different settings for GNN on Jetson Orin NX}}
	\setlength{\tabcolsep}{8pt}
	\renewcommand{\arraystretch}{1.1}
	\begin{tabular}{c|cccc|cccc|cccc}
		\hline
		\multirow{2}*{{Policy}} & \multicolumn{4}{c|}{{$N$ ($D = 0.1$ s, $B = 20$ MHz)}} & \multicolumn{4}{c|}{{$D$ ($N = 12$, $B = 20$ MHz)}} & \multicolumn{4}{c}{{$B$ ($N = 12$, $D = 0.1$ s)}} \\
		\multirow{2}*{} & 12 & 16 & 20 & 24 & 0.09 s & 0.10 s & 0.11 s & 0.12 s & 10 MHz & 15 MHz & 20 MHz & 25 MHz \\
		\hline
		Computing-only & 0.3569 & 0.4046 & 0.4361 & 0.4524 & 0.3791 & 0.3569 & 0.3316 & 0.3040 & 0.4479 & 0.3981 & 0.3569 & 0.2938          \\
		
		Memory-only & \textbf{0.3380} & \textbf{0.3824} & \textbf{0.4117} & \textbf{0.4268} & \textbf{0.3584} & \textbf{0.3380} & \textbf{0.3149} & \textbf{0.2896} & \textbf{0.4226} & \textbf{0.3763} & \textbf{0.3380} & 0.2795            \\
		
		Discrete-points & 0.3394 & 0.3840 & 0.4135 & 0.4287 & 0.3791 & 0.3394 & 0.3161 & 0.2907 & 0.4244 & 0.3779 & 0.3394 & 0.2805          \\
		\hline
		\textbf{Proposed (ours)} & \textbf{0.3380} & \textbf{0.3824} & \textbf{0.4117} & \textbf{0.4268} & \textbf{0.3584} & \textbf{0.3380} & \textbf{0.3149} & \textbf{0.2896} & \textbf{0.4226} & \textbf{0.3763} & \textbf{0.3380} & \textbf{0.2794}          \\
		\hline
	\end{tabular}
    \label{tab:V}
\end{table*}
The above results have shown the advantage of jointly optimizing memory frequency and computing frequency by comparing with several benchmark methods. To further verify the effect for other types of DNNs, we conduct ablation studies for LSTM on Jetson TX1 and GNN on Jetson Orin NX, and the results are shown in Table \ref{tab:IV} and Table \ref{tab:V}, respectively. In the computing-only policy, only the computing frequency is optimized while the memory frequency is fixed at the maximum value. In the memory-only policy, only the memory frequency is optimized while the computing frequency is fixed at the maximum value. The discrete-points policy searches over discrete memory and computing frequency configurations to obtain the appropriate combination, while the proposed method jointly optimizes both memory frequency and computing frequency in a continuous manner. The benefit of jointly optimizing memory frequency and computing frequency is highly dependent on the model architecture and hardware platform. For LSTM on Jetson TX1, the computing-only policy achieves energy consumption close to that of the proposed method, while the memory-only policy results in consistently higher energy consumption. This indicates that the optimal memory frequency is close to the maximum value. The fitting parameters also imply that the inference time of LSTM is more sensitive to memory frequency. For GNN on Jetson Orin NX, the memory-only policy significantly outperforms the computing-only policy and achieves almost the same performance as the proposed method, since the optimal computing frequency is pushed to the maximum value. It is observed that the proposed continuous frequency optimization and the discrete-points policy achieve close energy consumption. This is because when the available discrete frequency points are sufficiently dense and close to the continuous optimum, the discrete-points policy can approximate the proposed policy well. In fact, the proposed optimization framework can serve as a reference solution for practical implementations using discrete frequency points, where the continuous memory and computing frequency can be mapped to the nearest hardware-supported frequency points.

\subsubsection{Impact of Deadline and Bandwidth on Memory Frequency and Computing Frequency}
\begin{figure}[t]
	\captionsetup{font={small}}
	\centering
    \begin{subfigure}{0.8\linewidth}
        \centering
        \includegraphics[width=\linewidth]{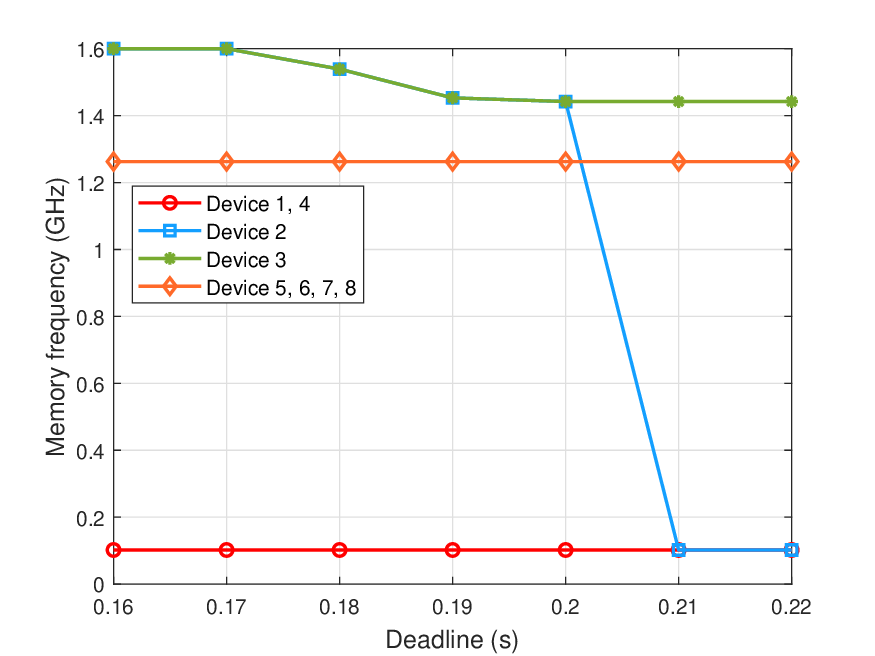}
        \caption{Impact of deadline on memory frequency}
        \label{fig:Deadline_MemFreq}
    \end{subfigure}
    \hfill
    \begin{subfigure}{0.8\linewidth}
        \centering
        \includegraphics[width=\linewidth]{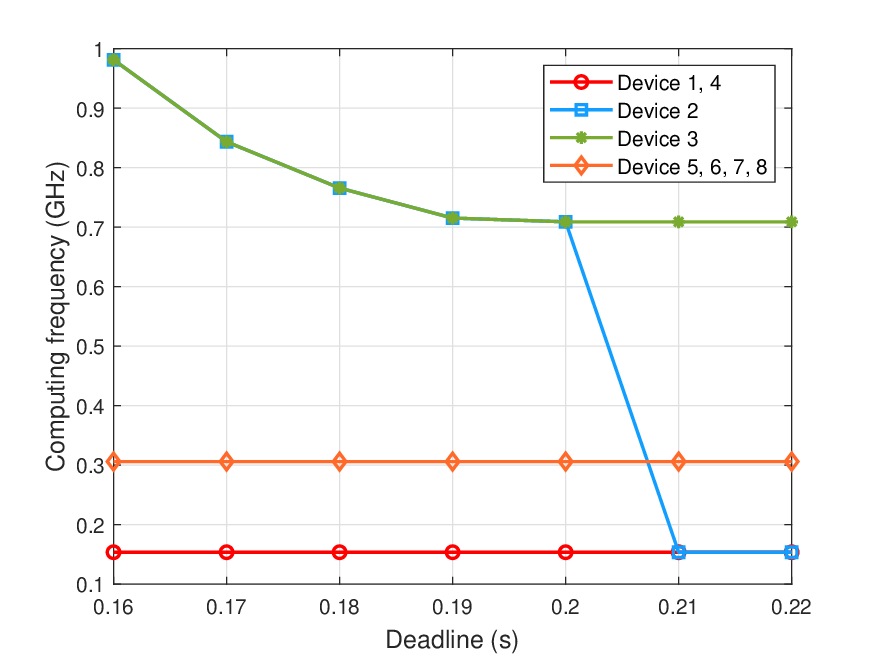}
        \caption{Impact of deadline on computing frequency}
        \label{fig:Deadline_ComFreq}
    \end{subfigure}
	\caption{Memory frequency and computing frequency solved by Algorithm \ref{alg:2} under different deadlines in a specific case.}\label{fig:Freq_Compare_Deadline}
	\vspace{-0.3cm}
\end{figure}
To further investigate the impact of deadline on memory frequency and computing frequency in the multi-device case, we consider a specific case with $N = 8$ mobile devices, where devices 1 to 4 use the Jetson TX1 configuration and the remaining devices use the Jetson Orin Nano configuration. VGG19 and ResNet152 are deployed on Jetson TX1 and Jetson Orin Nano, respectively, and the bandwidth is set to $B = 4$ MHz. We derive the memory frequency and computing frequency based on Algorithm \ref{alg:2}, and the results are shown in Fig. \ref{fig:Deadline_MemFreq} and Fig. \ref{fig:Deadline_ComFreq}. As the deadline increases, the memory frequencies and computing frequencies of devices 2 and 3 show a decreasing trend. When the deadline is set to $D_n = 0.21$ s, device 2 offloads its inference task, leading to the minimum memory frequency and computing frequency. On the one hand, increasing the deadline may lead to smaller memory frequencies and computing frequencies for energy-efficient DNN inference. On the other hand, devices have more opportunities to offload their inference tasks to reduce the energy consumption. With the limited bandwidth resources, devices 5 to 8 execute local inference. This is because the energy consumption of local inference on Jetson Orin Nano is smaller than that of Jetson TX1. In addition, the memory frequencies and computing frequencies of devices 5 to 8 do not decrease with the deadline, since the optimal energy consumption has been achieved. 

\begin{figure}[t]
	\captionsetup{font={small}}
	\centering
	\begin{subfigure}{0.8\linewidth}
        \centering
        \includegraphics[width=\linewidth]{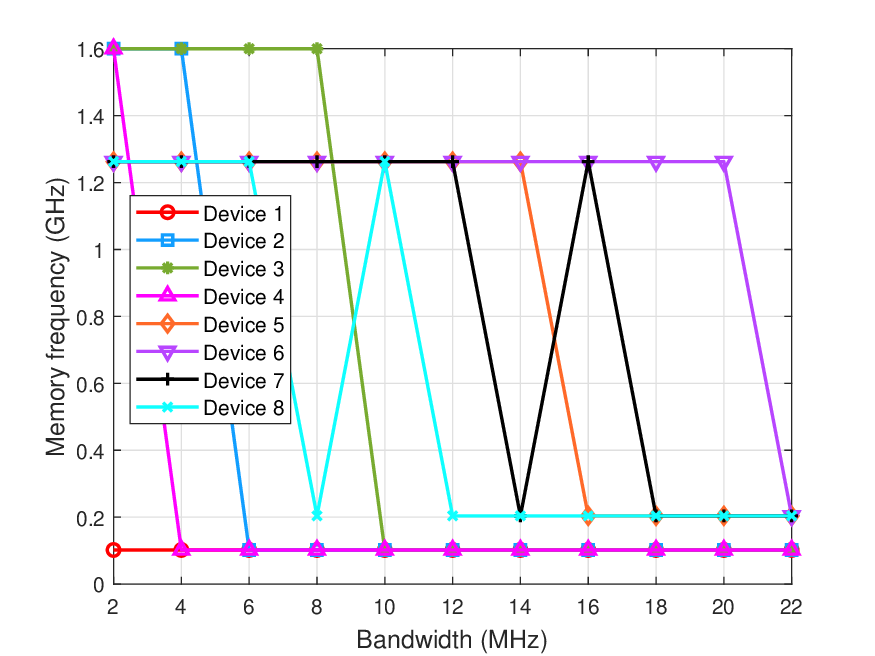}
        \caption{Impact of bandwidth on memory frequency}
        \label{fig:Bandwidth_MemFreq}
    \end{subfigure}
    \hfill
    \begin{subfigure}{0.8\linewidth}
        \centering
        \includegraphics[width=\linewidth]{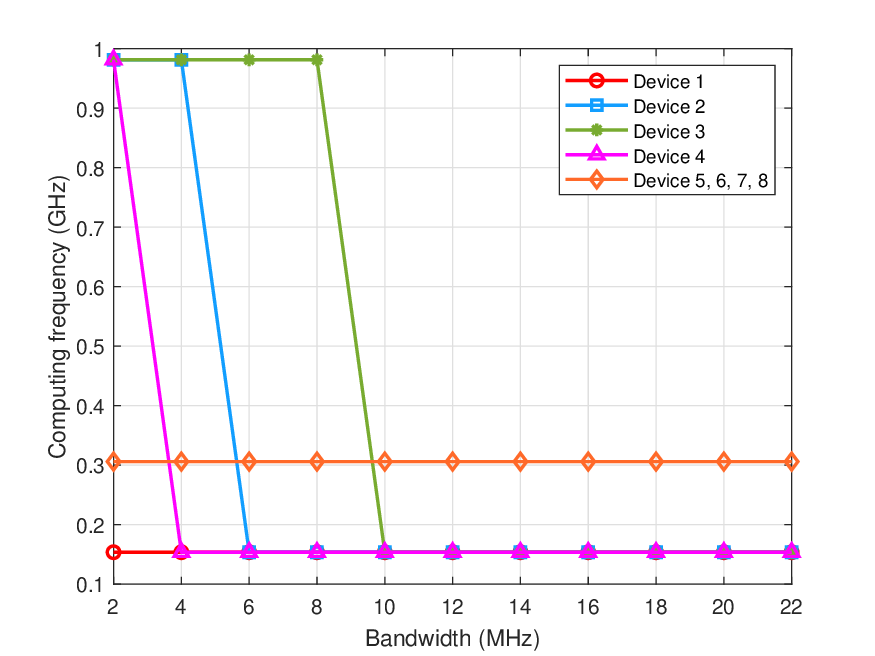}
        \caption{Impact of bandwidth on computing frequency}
        \label{fig:Bandwidth_ComFreq}
    \end{subfigure}
	\caption{Memory frequency and computing frequency solved by Algorithm \ref{alg:2} under different bandwidth resources in a specific case.}\label{fig:Freq_Compare_Bandwidth}
	\vspace{-0.3cm}
\end{figure}
Finally, we analyze the impact of bandwidth on memory frequency and computing frequency in the multi-device case. We consider the same scenario with $N = 8$ mobile devices, where the deadline is set to $D_n = 0.16$ s. The results are shown in Fig. \ref{fig:Bandwidth_MemFreq} and Fig. \ref{fig:Bandwidth_ComFreq}, respectively. As the bandwidth increases, there are more devices that can offload inference tasks, which leads to the decrease of memory frequency and computing frequency. When the bandwidth is set to $B = 22$ MHz, all mobile devices offload tasks to the edge node, and the memory frequency and computing frequency are adjusted to the minimum values. It can be observed that device 8 reduces the memory frequency to the minimum when $B = 8$ MHz, but increases the memory frequency when $B = 10$ MHz. A similar phenomenon can be observed for device 7 when $B = 14$ MHz and $B = 16$ MHz. This is because the system can adaptively optimize the bandwidth resource allocation to minimize the total energy consumption. In addition, the computing frequencies for devices 5 to 8 do not change with the bandwidth, since the computing frequency is scaled to the minimum even when devices execute local inference. For Jetson Orin Nano, the computing frequency range is relatively small, and the adjustment of memory frequency plays a more significant role in the inference time and energy consumption of DNN inference. In summary, Algorithm \ref{alg:2} can achieve an excellent balance between communication resources and memory/computing resources. 

\section{Conclusion}\label{sec:Conclusion}
In this paper, we investigated the joint optimization of memory frequency, computing frequency, bandwidth allocation, and offloading in the multi-device edge intelligence system. Motivated by the experiments that demonstrated the impact of memory frequency scaling on latency and energy consumption of DNN inference, we formulated an energy minimization problem under the deadline constraint and solved it through a decomposition approach. Our proposed algorithm integrated GP-based frequency optimization, bandwidth allocation with convex optimization, and a CD-based offloading policy optimization. Extensive evaluations based on real data were conducted under different parameter settings. The results demonstrate that our proposed algorithm can reduce the average energy consumption by at most $8.0\%$, $6.4\%$ and $16.1\%$ compared with the computing-only policy for ResNet152 on Jetson TX1, ViT-B/16 on Jetson Orin Nano and Swin-B on Jetson Orin NX, respectively, while the performance gaps from the optimal policy are within $6.4\%$, $5.6\%$ and $9.3\%$. 

Finally, several promising directions remain for future work. First, device mobility in practical edge intelligence systems may introduce channel uncertainty caused by fading. Therefore, extending the proposed framework to robust or chance-constrained resource allocation under channel uncertainty is an important direction. Second, when multiple devices share the MEC server, resource contention and queueing delay may make the edge inference time variable rather than constant. This motivates the joint optimization of computing/memory resources at both mobile devices and the MEC server. Third, the proposed framework can be further extended to collaborative device-edge inference with DNN partitioning, where the latency and energy consumption of memory/computing frequency scaling should be characterized at the DNN block level. The corresponding optimization problem would need to jointly consider DNN partitioning, memory/computing frequency scaling, and wireless bandwidth allocation.

\baselineskip=18pt
\bibliographystyle{IEEEtran}

\ifCLASSOPTIONcaptionsoff
  \newpage
\fi

\begin{IEEEbiography}[{\includegraphics[width=1in,height=1.25in,clip,keepaspectratio]{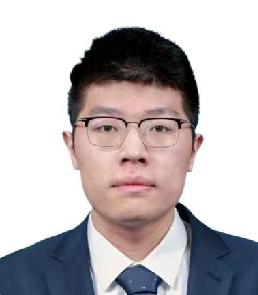}}]{Yunchu Han} (S'24) received the B.E. degree from the Department of Electronic Engineering, Tsinghua University in 2023. He is currently pursuing the Ph.D. degree with the Network Integration for Ubiquitous Linkage and Broadband Laboratory, Department of Electronic Engineering, Tsinghua University. His research interests include green edge intelligence, efficient inference, mobile edge computing and wireless communication.
\end{IEEEbiography}

\vspace{-5 mm}

\begin{IEEEbiography}[{\includegraphics[width=1in,height=1.25in,clip,keepaspectratio]{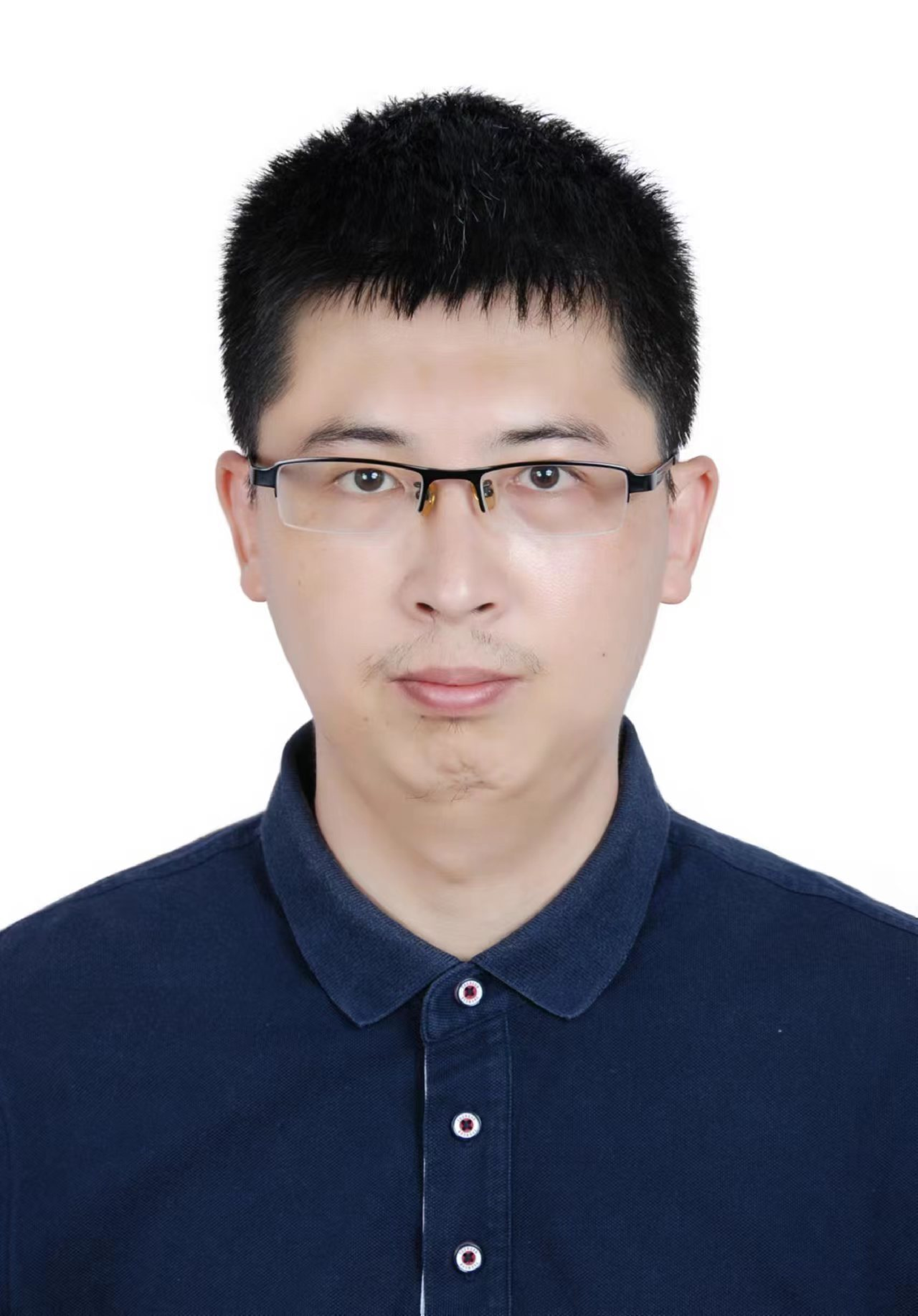}}]{Zhaojun Nan} (S'20-M'23) received the B.S. degree in automation from Jiamusi University, Jiamusi, China, in 2009, the M.S. degree in navigation, guidance and control from Harbin Engineering University, Harbin, China, in 2012, and the Ph.D. degree in information and communication engineering from Chongqing University, Chongqing, China, in 2022. From 2012 to 2017, he worked for Tianjin 712 Communication $\&$ Broadcasting Co., Ltd., Tianjin, China. From 2022 to 2025, he was a Postdoctoral Researcher with the Department of Electronic Engineering, Tsinghua University, Beijing, China. He is currently with the School of Electronics and Internet of Things, Chongqing Polytechnic University of Electronic Technology, Chongqing, China. His research interests include mobile edge intelligence, vehicular networks and autonomous driving, and green wireless communications. He has served as a TPC member for IEEE Globecom, ICC and WCNC.
\end{IEEEbiography}

\vspace{-5 mm}

\begin{IEEEbiography}[{\includegraphics[width=1in,height=1.25in,clip,keepaspectratio]{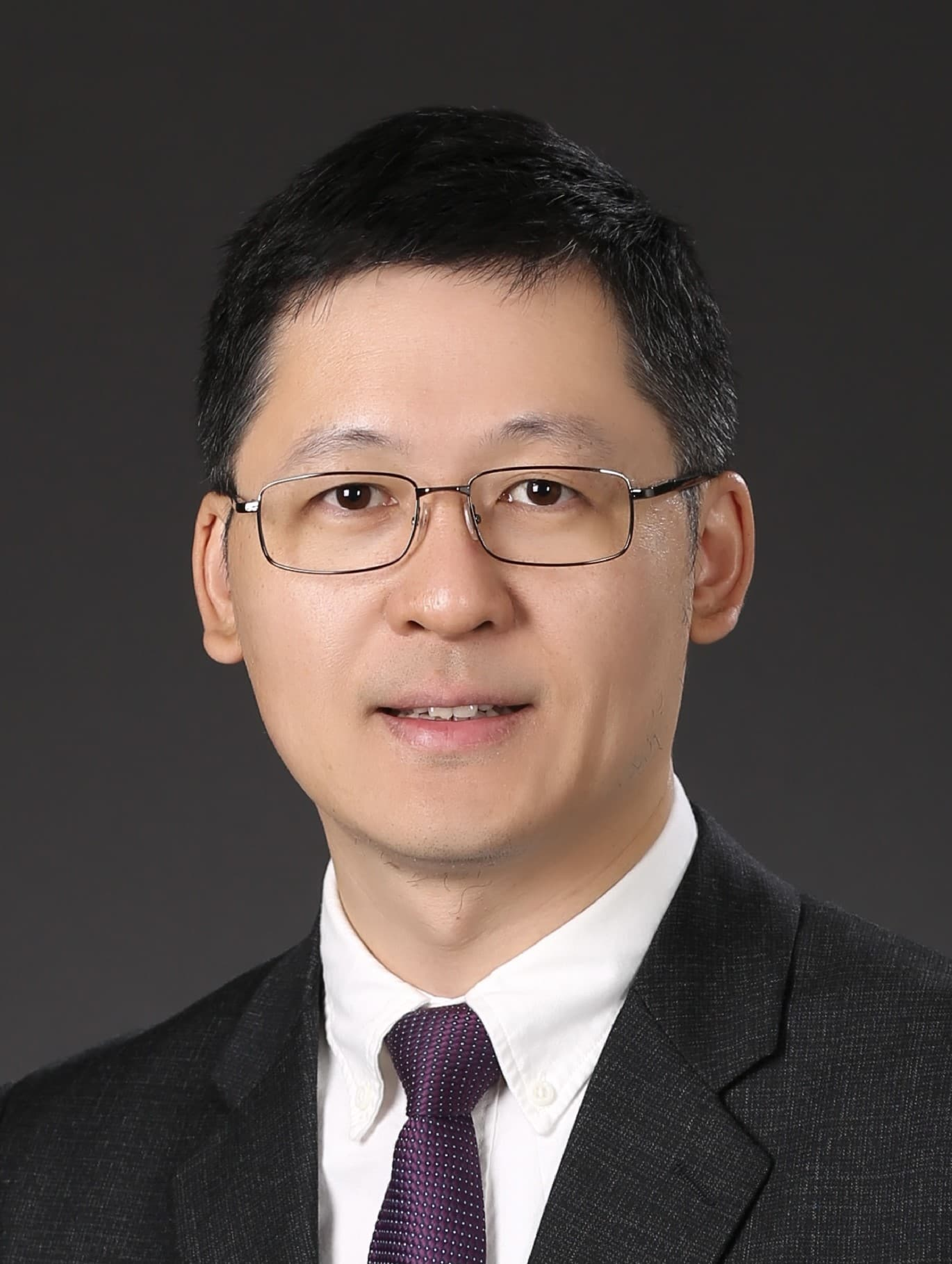}}]{Sheng Zhou} (S'06-M'12) received the B.E. and Ph.D. degrees in electronic engineering from Tsinghua University, Beijing, China, in 2005 and 2011, respectively. In 2010, he was a Visiting Student with the Wireless System Lab, Department of Electrical Engineering, Stanford University, Stanford, CA, USA. From 2014 to 2015, he was a Visiting Researcher with Central Research Lab, Hitachi Ltd., Tokyo, Japan. He is currently an Associate Professor with the Department of Electronic Engineering, Tsinghua University, Beijing, China. His research interests include cross-layer design for multiple antenna systems, mobile edge computing, vehicular networks, and green wireless communications. He was the recipient of the IEEE ComSoc Asia-Pacific Board Outstanding Young Researcher Award in 2017, and IEEE ComSoc Wireless Communications Technical Committee Outstanding Young Researcher Award in 2020.
\end{IEEEbiography}

\vspace{-5mm}

\begin{IEEEbiography}[{\includegraphics[width=1in,height=1.25in,clip,keepaspectratio]{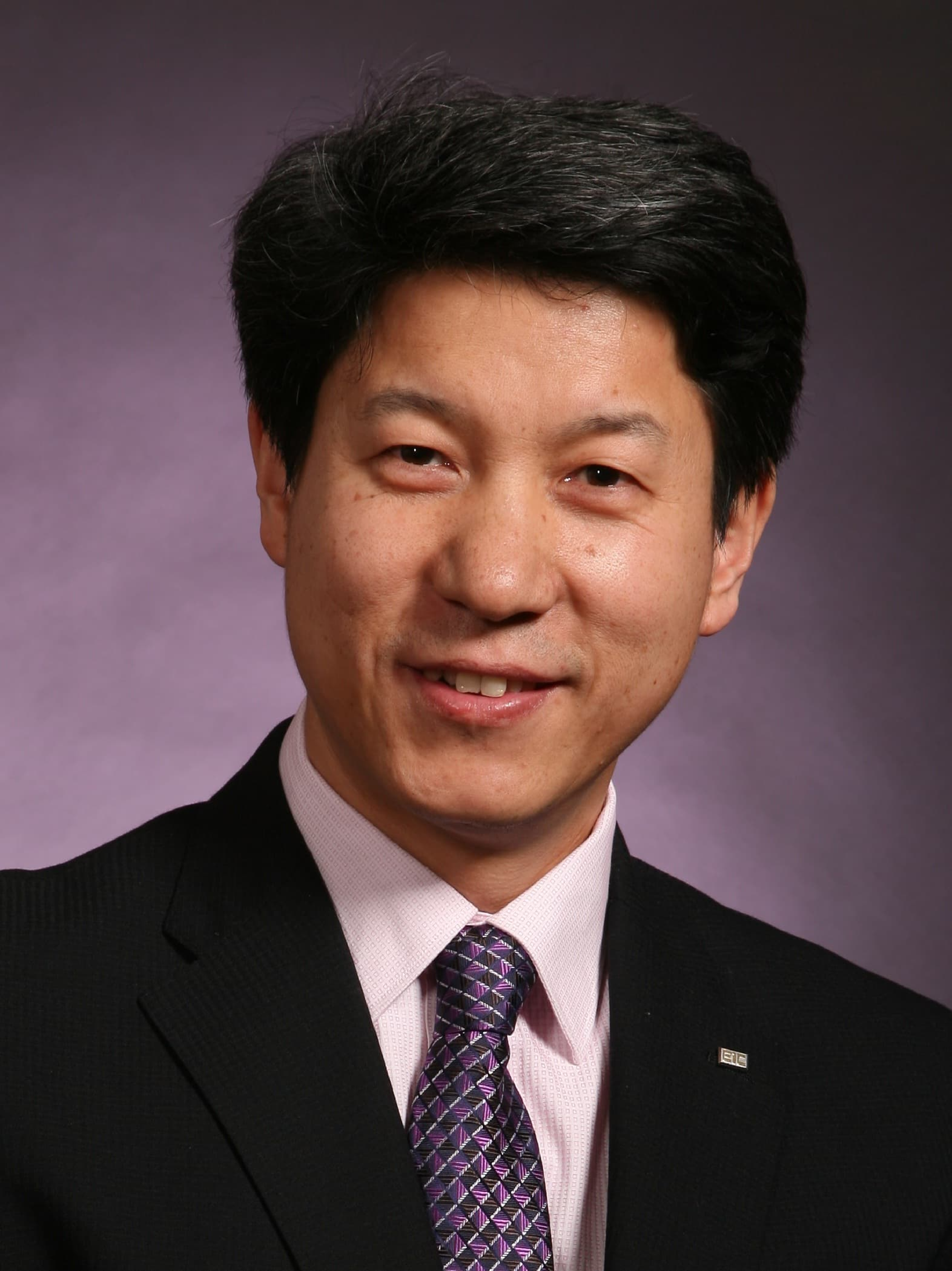}}]{Zhisheng Niu} (M'98-SM'99-F'12) graduated from Beijing Jiaotong University, China, in 1985, and got his M.E. and D.E. degrees from Toyohashi University of Technology, Japan, in 1989 and 1992, respectively.  From 1992 to 1994, he worked for Fujitsu Laboratories Ltd., Japan, and in 1994 joined with Tsinghua University, Beijing, China, where he is now a Professor at the Department of Electronic Engineering. His major research interests include queueing theory, traffic engineering, mobile Internet, radio resource management of wireless networks, and green communication and networks.

Dr. Niu has been serving IEEE Communications Society since 2000 as Chair of Beijing Chapter (2000-2008), Director of Asia-Pacific Board (2008-2009), Director for Conference Publications (2010-2011), Chair of Emerging Technologies Committee (2014-2015), Director for Online Contents (2018-2019) and the Editor-in-Chief of \textsc{IEEE Transactions on Green Communications and Networking} (2020-2022). He received the Best Paper Award of Asia-Pacific Board in 2013, Distinguished Technical Achievement Recognition Award of Green Communications and Computing Technical Committee in 2018, and Harold Sobol Award for Exemplary Service to Meetings \& Conferences in 2019, all from the IEEE Communications Society. He was selected as a distinguished lecturer of IEEE Communications Society (2012-2015) as well as IEEE Vehicular Technologies Society (2014-2018). He is a fellow of both IEEE and IEICE.
\end{IEEEbiography}

\end{document}